%% file: main.tex
\documentclass[11pt]{article}

\usepackage[final]{acl}

\usepackage{times}
\usepackage{latexsym}

\usepackage[T1]{fontenc}

\usepackage[utf8]{inputenc}

\usepackage{microtype}

\usepackage{inconsolata}

\usepackage{hyperref}       
\usepackage{url}            
\usepackage{booktabs}       
\usepackage{amsfonts}       
\usepackage{nicefrac}       
\usepackage{xcolor}         
\usepackage{algorithm}      %
\usepackage{algpseudocode}  %
\usepackage{amsmath}        %
\usepackage{amssymb}

\usepackage[textsize=tiny]{todonotes}
\usepackage{xcolor} %
\usepackage{enumitem}
\usepackage{amsmath} 
\usepackage{amssymb}        %
\usepackage{hyperref}       %
\usepackage{array}
\usepackage{wrapfig}
\usepackage{makecell}
\usepackage{amsmath}
\usepackage{amssymb}
\usepackage{mathtools}
\usepackage{amsthm}
\usepackage{graphicx}
\usepackage{subfigure}
\usepackage{booktabs} 
\usepackage{xcolor}
\usepackage{subcaption} 

\usepackage{fancyvrb}

\title{Efficient Multi-Agent System Training with Data Influence-\\Oriented Tree Search}

\author{Wentao Shi\thanks{Work done during the internship at Carnegie Mellon University.}$^{\spadesuit}$, 
Zichun Yu$^{\clubsuit}$,
Fuli Feng$^{\spadesuit}$,
Xiangnan He$^{\spadesuit}$,
Chenyan Xiong\thanks{Corresponding Author}$^{\clubsuit}$,
\\
$^{\spadesuit}$University of Science and Technology of China \quad
$^{\clubsuit}$Carnegie Mellon University\\
\texttt{shiwentao123@mail.ustc.edu.cn, \{fulifeng93, xiangnanhe\}@gmail.com}, \\ \{zichunyu, cx\}@andrew.cmu.edu
}

\begin{document}
\maketitle

\begin{abstract}
\input{section/0_abstract}
\end{abstract}

\input{section/1_Introduction}

\input{section/4_Methods}

\input{section/5_Setup}

\input{section/6_Experiments}

\input{section/7_Conclusion}

\input{section/8_Limitation}


\bibliography{custom}

\appendix
\include{section/Appendix}

\end{document}

%% file: section/0_abstract.tex

Large Language Model (LLM) based multi-agent systems (MAS) show strong potential for tackling complex tasks through collaborative intelligence. Monte Carlo Tree Search (MCTS) based methods provide promising approaches for enhancing MAS self-training by generating synthetic data, using Q-values to estimate agent contributions. However, relying solely on Q-values may misalign with the goal of selecting data most beneficial for MAS improvement. To address this discrepancy, we propose \textbf{D}ata \textbf{I}nfluence-oriented \textbf{T}ree \textbf{S}earch (\textbf{DITS}), a novel framework that incorporates influence scores to guide both tree search and data selection in data synthesis. By leveraging influence scores, we effectively identify the most impactful data for MAS improvement, thereby enhancing model performance. Furthermore, we derive a novel influence score estimation method tailored for non-differentiable metrics, significantly reducing computational overhead by calculating performance changes on the validation set. Extensive experiments on three different multi-agent tasks demonstrate the robustness and effectiveness of the proposed methods. Notably, our findings reveal that allocating more resources to estimate influence scores, rather than Q-values, during data synthesis can more effectively and efficiently enhance model training. The code is available at \href{https://github.com/swt-user/DITS}{https://github.com/swt-user/DITS}.

%% file: section/1_Introduction.tex
\section{Introduction}
\label{section:introduction}

LLM based agents have recently achieved remarkable success across a wide range of tasks~\citep{202412.2294, Wang_2024, xi2023risepotentiallargelanguage, zhang2024largelanguagemodelbrainedgui}. Leveraging the advanced natural language understanding and reasoning capabilities of LLMs~\citep{DBLP:journals/corr/abs-2303-08774, DBLP:conf/nips/Wei0SBIXCLZ22}, these agents are able to dynamically interact with complex tools and environments to accomplish various tasks~\citep{ DBLP:journals/corr/abs-2310-05915, DBLP:conf/iclr/YaoZYDSN023}. Nevertheless, individual agents often face significant limitations when confronted with complex tasks~\citep{DBLP:conf/emnlp/ShiYWWF24}. In such scenarios, the multi-agent system (MAS) (e.g., MetaGPT~\citep{DBLP:conf/iclr/HongZCZCWZWYLZR24}, AutoGen~\citep{DBLP:journals/corr/abs-2308-08155}, Camel~\citep{DBLP:conf/nips/LiHIKG23}) involving multiple specialized agents, with strategic task allocation and division of labor, becomes crucial for achieving optimal outcomes~\citep{DBLP:conf/ijcai/GuoCWCPCW024}. However, optimizing the collective performance of LLM-based MAS as a cohesive unit and obtaining reward signals for each agent in the MAS still remain challenging problems~\citep{DBLP:journals/corr/abs-2410-08115}.


\begin{figure}
    \centering
    \subfigure[Data Distribution]{\includegraphics[width=0.49\linewidth]{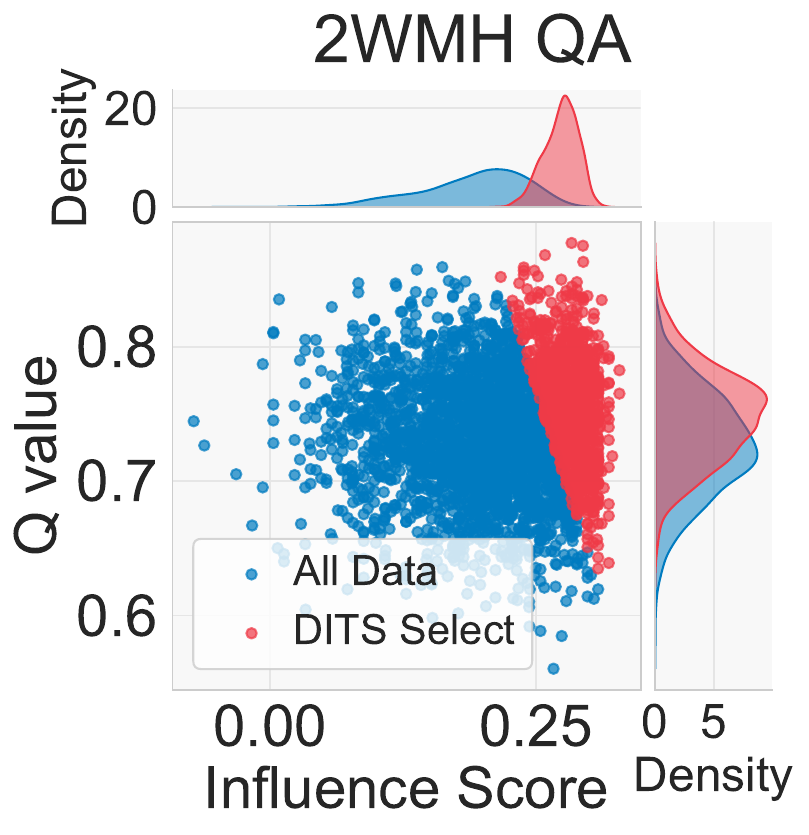}}
    \subfigure[Synthesis Time Scaling]{\includegraphics[width=0.49\linewidth]{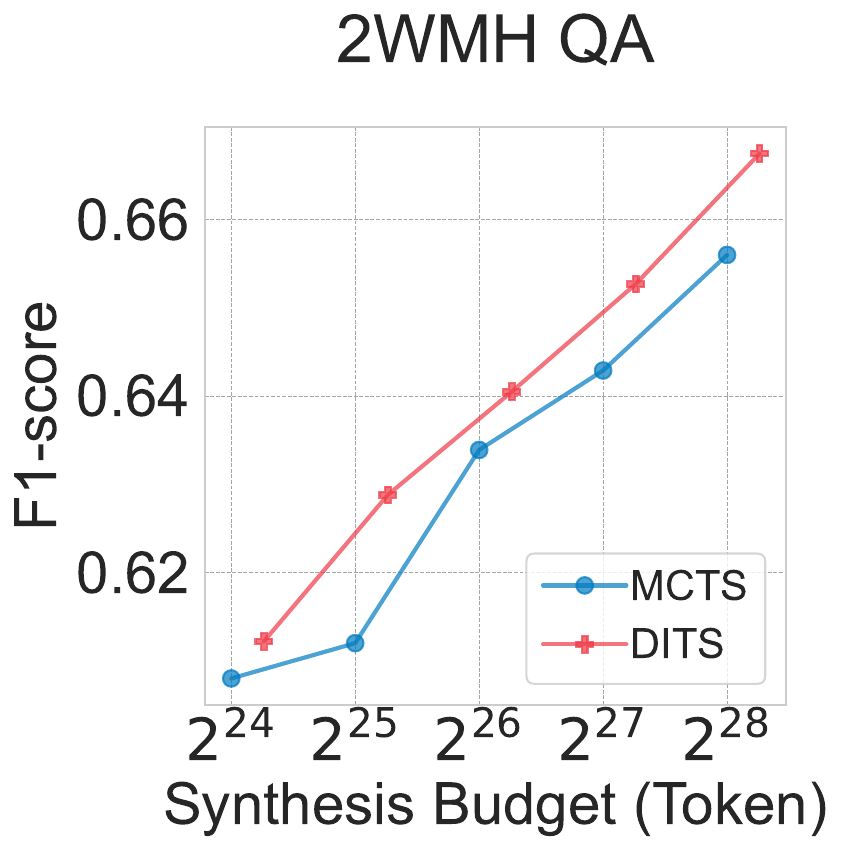}}
    \caption{(a) The scatter plot and density plots of Q-values and influence scores for synthetic data. The top 30\% of the data selected using DITS is highlighted in red. (b) Performance trends with different data synthesis budgets (Tokens).}
    \vspace{-10pt}
    \label{fig:scaling}
\end{figure}

To tackle this challenge, leveraging synthetic data for self-training emerges as a highly promising direction. Monte Carlo Tree Search (MCTS)~\citep{guan2025rstarmathsmallllmsmaster, li2025enhancingreasoningprocesssupervision} based method offers a promising approach for synthetic data generation, capable of estimating individual agent contributions through Q-value~\citep{DBLP:journals/corr/abs-2410-08115}. They collect fine-grained preference pairs, encouraging high-Q-value actions while suppressing low-Q-value actions via Direct Preference Optimization (DPO)~\citep{DBLP:conf/nips/RafailovSMMEF23}. 
Despite its potential, the current tree search strategy is primarily adapted from the inference phase, inheriting its inherent characteristics, which rely on Q-values to identify informative data. This reliance misaligns with the data synthesis objective, which focuses on generating data that better facilitates model training. The empirical results presented in Figure~\ref{fig:scaling} (a) also demonstrate that actions associated with higher Q-values do not always contribute significantly to the improvement of model performance, where the influence score serves as a metric to quantify the utility of data in enhancing performance.

To address this issue, we propose \textbf{D}ata \textbf{I}nfluence-oriented \textbf{T}ree \textbf{S}earch (\textbf{DITS}), a novel framework that optimizes MAS through iterative synthetic data generation guided by influence-aware tree search. Our approach combines MCTS for MAS trajectory simulation with a data influence mechanism that prioritizes training samples based on their expected contribution to model improvement, rather than relying solely on traditional Q-value estimates.
The influence score quantifies how training data impacts model outputs, helping identify data points that most improve performance. While traditional methods rely on training loss as a performance metric, this is less effective for DPO loss due to its weak correlation with downstream performance~\citep{DBLP:journals/corr/abs-2406-02900, DBLP:journals/corr/abs-2410-11677}. Hence, we redefine the influence score based on the changes in non-differentiable metrics on the validation set and derive a novel estimation method. Our method circumvents computationally intensive gradient computations across large-scale parameters that are required in traditional approaches. 

We validate our approach on seven datasets across three multi-agent tasks: Information Exchange, Debate~\citep{DBLP:journals/corr/abs-2410-08115}, and DeepSearch~\citep{Li2025WebThinker}. We observe that high Q-value data may reduce the diversity of the model's responses and contribute little to improving model performance. Incorporating data influence is crucial for data synthesis and selection. Our method outperforms state-of-the-art multi-agent optimization techniques, achieving an average improvement of 2.7\% in single-round iterations, a 2.5\% performance enhancement in multi-round iterations for the Information Exchange task, and 2.6\% performance improvement for the DeepSearch task. Within the same data synthesis budget, our method surpasses traditional approaches, delivering more efficient scaling of synthesis computation, as shown in Figure~\ref{fig:scaling} (b) and in Section~\ref{subsection:synthesis time scaling}.

We summarize the contributions as follows: 
\begin{itemize}[leftmargin=*]
    \item We propose DITS, a novel framework that employs influence scores to guide tree search and data selection. This enables the prioritized selection of preference pairs that contribute more significantly to performance improvement.
    
    \item We derive the influence score estimation method for non-differentiable metrics. This approach substantially reduces computational overhead through inference computation, enabling more efficient synthesis time scaling.
    \item We achieve state-of-the-art performance across multiple multi-agent tasks and demonstrate that the framework's capability can be improved through iterative rounds of data synthesis.
\end{itemize}

%% file: section/4_Methods.tex
\section{Method}
\label{section:method}
In this section, we first formalize the multi-agent task and MCTS-based data synthesis (§~\ref{subsection:multi-agent-data-synthesis}), then introduce the data influence-oriented data selection
(§~\ref{subsection:hybrid data selection}), and finally present the iterative data synthesis process (§~\ref{subsection:iterative data synthesis}).

\subsection{Multi-Agent Training Data Synthesis} \label{subsection:multi-agent-data-synthesis}
Training effective MAS requires high-quality data that reflects complex agent interactions, but collecting such data in the real world is costly and time-consuming. To overcome this, we utilize MCTS to simulate interactions and automatically produce preference-labeled training data.

\begin{figure*}
    \centering
    \includegraphics[width=0.98\linewidth]{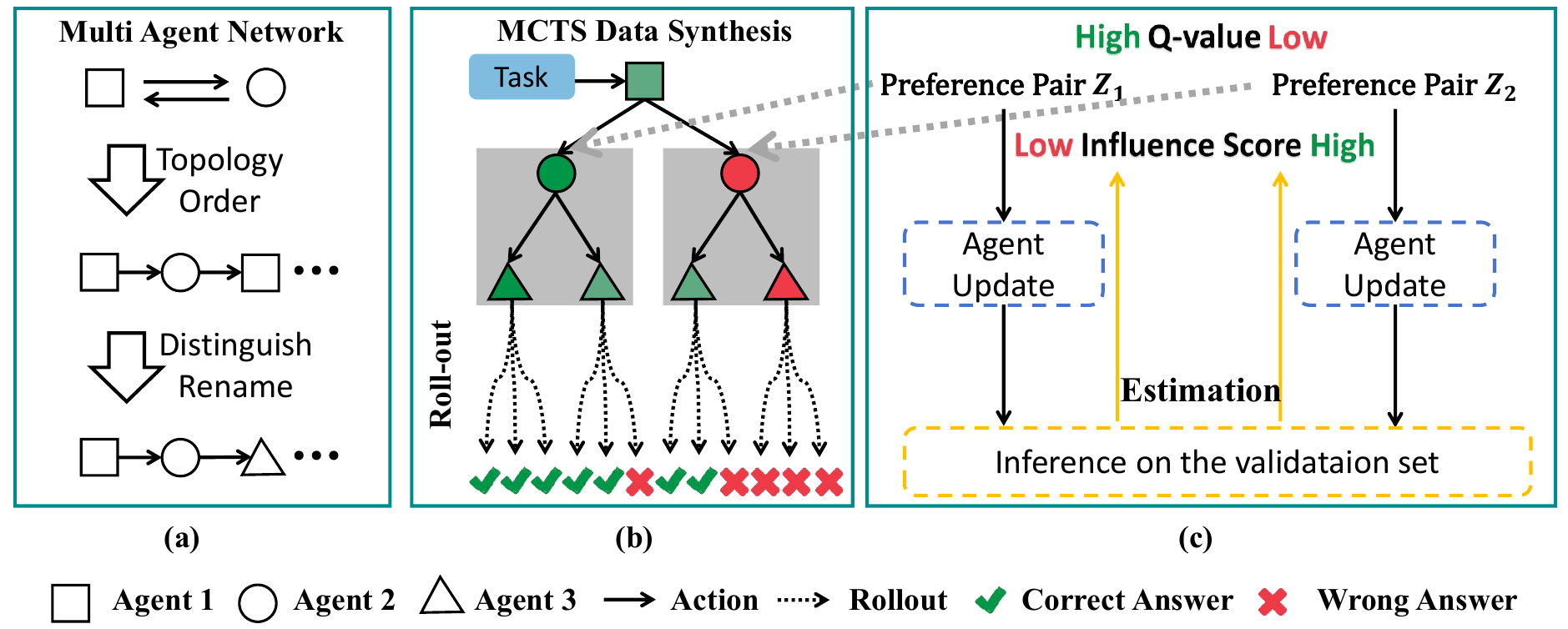}
    \caption{Overview of our method. (a) illustrates the traversal of a cyclic agent network in topological order. We introduce virtual agents to distinguish the same agent in the traversal. (b) showcases the application of MCTS to generate synthetic multi-agent training data, where the color of each agent represents the magnitude of the node's Q-value. (c) depicts the computation process of influence scores for a non-differentiable metric, highlighting that data points with high Q-values may correspond to low influence scores.} 
    \label{fig:framework}
    
\end{figure*}

In this work, we model the topology structure for multi-agent collaboration as a directed graph. Concretely, we denote a feasible topology as $\mathcal{G}=(\mathcal{V}, \mathcal{E})$, as demonstrated in Figure~\ref{fig:framework} (a). It is worth noting that such graph structures can be static or dynamic, with the dynamic variant allowing agents to govern the information flow in an adaptive manner. We allow the presence of cycles in the graph, indicating that multiple rounds of information exchange are permitted among agents $A$. We assume that our agent network can be linearly traversed in topological order $A_1\oplus A_2 \oplus \cdots \oplus A_M$~\cite{Bondy1976, book:gross:2005, qian2024}, where $A_m \in \mathcal{V}$. Different $A_m$ may represent the same agent being visited at different time steps. For clarity and convenience, we use different symbols to distinguish them.


In this way, we could utilize MCTS to synthesize training data for MAS. We mainly follow the configuration in Optima~\cite{DBLP:journals/corr/abs-2410-08115} and construct the tree as follows: As shown in Figure~\ref{fig:framework} (b), the synthesis tree begins with a specific task instruction $p$. 

\textbf{Selection}: We select a node $n$ to
expand from the candidate node set, where a node $n=(s,a)$ refers to an agent $A_m$ in state $s$ that takes action $a$. We use the edit distance to filter out nodes that are similar to expanded nodes to obtain the candidate node set.
\begin{equation}
    N_{\text{cand}} = \{n_j| n_i \in N_{\text{exp}}, n_j\in N_{\text{all}}, S_{i,j} \geq 0.25 \},
\end{equation}
where $S_{i,j}=\frac{\text{edit\_distance}(n_i, n_j)}{\max (|n_i|,|n_j|)}$ and $\text{edit\_distance}(n_i, n_j)$ represents the edit distance between the action strings of two nodes. $N_\text{all}$ and $N_{\text{exp}}$ denotes the whole node set and expanded node set. Then we select a node for the candidate set $N_{\text{cand}}$ based on softmax distribution of Q-values.
\begin{equation}
n\sim \text{Softmax}(\{Q(n)\}_{n\in N_{\text{cand}}}),
\end{equation}
where $Q(n) = Q(s, a)$ and the softmax distribution balances exploration and exploitation.

\textbf{Expansion} For each selected node $n$, we denote the new state as $s'=\text{Trans}(s, a)$, where $\text{Trans}(\cdot)$ is the transit function determined by the environment. Then we sample $d$ actions from agents $A_{m+1}$: 
\begin{equation}
    \{a_1', \cdots, a_d' \} \sim A_{m+1}(s').
\end{equation}

\textbf{Simulation} For each generated action $a_i'$, we simulate the agent interaction $\tau_i$ until termination.
\begin{equation}
    \tau_i = \text{Simulation}(A_{m+2}, \cdots, A_{M}, s', a_i').
\end{equation}
Meanwhile, we construct all $(s, a)$ pairs in the trajectory as new nodes and add them to $N_{\text{all}}$.

\textbf{Backpropagation} Once a trajectory $\tau$ is completed, we can obtain the trajectory reward $R(\tau)$ detailed in Appendix~\ref{appendix:method_details}. We update the Q-value of nodes with the average rewards from the trajectories set containing the node.
\begin{equation}
    Q(n) = Q(s,a) = \sum_{\tau \in \mathcal{T}(n)} \frac{1}{|\mathcal{T}(n)|} R(\tau),
\end{equation}
where $\mathcal{T}(n)$ denotes the trajectory set containing the node $n$. Additionally, due to the complex interactions among multiple agents, the Q-value estimates obtained from $d$ rollouts may be inaccurate. Allocating more inference budget in the data synthesis phase may improve the quality of the generated data and enhance the system’s performance.

We repeat the above process $k$ times and finish the generation process. Then we can construct paired action preferences for agent $A_i$ at state $s$ by selecting the action $a_i^h$ with the highest Q-value and the action $a_i^l$ with the lowest Q-value to form the preference data:
\begin{equation}
    z  = \left(s, a_i^{h}, a_i^{l}\right).
\end{equation}
To update the parameter of agent $A_i$, we utilize the Direct Preference Optimization (DPO) loss to encourage the model to prioritize responses that align with preferences $a_i^{h}$ over less preferred $a_i^{l}$.
\begin{equation}
\begin{aligned}
    \mathcal{L}_{DPO} = \mathbb{E}_{z} \bigg[ 
    -\log \sigma \bigg( \beta \bigg[
        \log \frac{\pi_\theta(a_i^{h} \mid s)}{\pi_{\text{ref}}(a_i^{h} \mid s)} 
        \\
        - \log \frac{\pi_\theta(a_i^{l} \mid s)}{\pi_{\text{ref}}(a_i^{l} \mid s)}
    \bigg] \bigg) \bigg],
\end{aligned}
\end{equation}

where $\sigma(\cdot)$ denotes the sigmoid function and $\pi_{\text{ref}}$ represents the reference model, \textit{i.e.} the SFT model.

\subsection{Data Influence-Oriented Data Selection}
\label{subsection:hybrid data selection}


While improving the accuracy of Q-value estimation can enhance data quality to some extent, it is both highly inefficient and suboptimal. During the training phase, the primary goal of synthetic data is to maximize its contribution to model performance improvement, rather than ensuring the data is correct. Figure~\ref{fig:framework} (c) reveals an important insight: while the data pair $z_1$ achieves higher Q-values, the data pair $z_2$ demonstrates greater practical impact on system performance. This suggests that absolute Q-values may not fully capture data pair's true contribution.


Hence, in this paper, we introduce the influence score $\mathcal{I}$ to quantify the impact of data on the current agent's performance. 
The influence score $\mathcal{I}$ was developed to measure the difference in loss when a data point is assigned a higher weight in the training dataset.  Suppose the agent $A$ is parameterized by $\theta$. We denote the optimal parameters learned by minimizing the training loss $\mathcal{L}_{\text{tr}}$ on the dataset $\mathcal{D}_{\text{tr}}$, with a data point $z_i$ assigned an additional weight of $\epsilon$, as:
\begin{equation}
    \theta_{\epsilon, z_i}^* = \underset{\theta}{\arg\min} \sum_{z_j\in \mathcal{D}_{\text{tr}}} \frac{1}{|\mathcal{D}_{\text{tr}}|} \mathcal{L}_{\text{tr}}(z_j, \theta) + \epsilon \mathcal{L}_{\text{tr}}(z_i, \theta).
\end{equation}
Under standard assumptions, such as the twice-differentiability and strong convexity of the loss function $\mathcal{L}_{\text{tr}}$, the influence function can be derived via the chain rule of the derivatives~\citep{DBLP:conf/icml/KohL17}.
However, the DPO loss does not effectively align with downstream task performance. Our experiments reveal a weak correlation (less than 0.2) between the DPO loss and performance metrics $\mathcal{F}$ such as F1-score or Accuracy on the validation set. This observation is consistent with findings reported in~\citep{DBLP:journals/corr/abs-2406-02900, DBLP:journals/corr/abs-2410-11677}. This indicates that we must redefine the influence score using the changes of non-differentiable performance metrics on the validation set.
\begin{equation}
\mathcal{I}_{\mathcal{F}_{\text{val}}}(z_i, \mathcal{D}_{\text{val}}) := \frac{\mathcal{F}_{\text{val}}(z_i, \theta_{\epsilon,z_i}^*) - \mathcal{F}_{\text{val}}(z_i, \theta^*)}{\epsilon},
\end{equation}

where $\theta^*=\theta_{\epsilon,z_i}^*|_{\epsilon=0}$. Due to non-differentiable metric  $\mathcal{F}_{\text{val}}$, the influence function cannot be derived using gradients. Instead, we use the finite difference method combined with parameter perturbation to approximate the rate of change. The perturbed optimal parameter $\theta_{\epsilon,z_i}^*$ can be rewritten as:
\begin{equation}
    \theta_{\epsilon,z_i}^* = \theta^* + \epsilon\Delta \theta + o(\epsilon),
\end{equation}
where $\Delta\theta$ represents the direction of parameter change. Following~\cite{yu2024mates}, the direct is typically driven by the gradient of the training loss.
\begin{equation}
    \Delta \theta \propto -\nabla_\theta \mathcal{L}_{\text{tr}}(z_i, \theta^*).
\end{equation}
Since the parameter update is dominated by the training loss gradient, we adopt a one-step gradient descent update:
\begin{equation}
    \theta_{\epsilon,z_i}^* \approx \theta^* - \eta\epsilon\nabla_\theta \mathcal{L}_{\text{tr}}(z_i,\theta^*),
\end{equation}
where $\eta$ is the learning rate, and $\epsilon$ is a very small perturbation strength. Combining the finite difference and parameter update, the influence function is approximated as:
\begin{equation}
\begin{split}
    &\mathcal{I}_{\mathcal{F}_{\text{val}}}(z_i, \mathcal{D}_{\text{val}}, \theta^*) \approx  \frac{1}{\epsilon} \left[ \right. \\
    &\left. \mathcal{F}_{\text{val}}(z_i,  \theta^* - \eta\epsilon\nabla_\theta L_{\text{tr}}(z_i,\theta^*)) 
      - \mathcal{F}_{\text{val}}(z_i,\theta^*) \right].
\end{split}
\label{eq:influence_estimation}
\end{equation}

Following~\cite{DBLP:conf/icml/KohL17}, we theoretically illustrate that selecting data points with the highest influence scores maximizes the model's validation performance (see Appendix~\ref{appendix: theoretical analysis} for details). 
Finally, our selection strategy combines Q-values and influence scores to effectively identify the highest-quality pair data: 
\begin{equation}
\begin{split}
    H(z_i) = \mathcal{I}_{\mathcal{F}_{\text{val}}}(z_i, \mathcal{D}_{\text{val}}, \theta) + \gamma \cdot Q(s, a_i^h) , 
    \label{equation:hybrid score}
\end{split}
\end{equation}
where $\theta$ denotes the current parameters of agent $A_m$. Finally, after filtering out low-quality data as described in~\cite{DBLP:journals/corr/abs-2410-08115}, synthetic data are ranked based on the scores, and the Top $\alpha$ are selected to construct the training dataset $\mathcal{D}_{\text{tr}}$.

\subsection{Iterative Data Synthesis}
\label{subsection:iterative data synthesis}
In addition to utilizing the current model for data synthesis, we propose an iterative refinement approach to generate higher-quality data. By continuously training and enhancing the model, its capabilities improve, enabling the generation of more valuable synthetic data in subsequent iterations. At iteration $t$, we generate the training dataset $\mathcal{D}_{\text{tr}}^t$ based on the parameters $\theta_{t-1}$ and train a new model from the initial model using $\mathcal{D}_{\text{tr}}^t$. The corresponding pseudocode can be found in Algorithm~\ref{alg:DITS-isft-dpo}.

%% file: section/5_Setup.tex
\input{table/main_table}

\section{Experimental Setup}
\label{section:setup}

In this section, we will introduce the datasets, metrics, and baseline methods in the experiments.

\textbf{Dataset}
To validate the collaborative and task allocation capabilities of MAS,  
we evaluate our framework DITS in three multi-agent settings: two static scenarios—\textbf{Information Exchange} and \textbf{Debate}—and one dynamic scenario, \textbf{DeepSearch}. The information exchange setting includes HotpotQA~\citep{DBLP:conf/emnlp/Yang0ZBCSM18}, 2WikiMultiHopQA (2WMH QA)~\citep{DBLP:conf/coling/HoNSA20}, TriviaQA~\citep{DBLP:conf/acl/JoshiCWZ17}, and CBT~\citep{DBLP:journals/corr/HillBCW15}. 
The debate setting includes ARC's challenge set (ARC-C)~\citep{DBLP:journals/corr/abs-2102-03315} and MMLU~\citep{DBLP:conf/iclr/HendrycksBBZMSS21}. 
The DeepSearch setting includes WebWalker~\citep{wu2025webwalker}. We use 0-shot for all benchmarks. More details can be found in Appendix~\ref{appendix:dataset}.

\textbf{Metrics}
Following~\citet{DBLP:journals/corr/abs-2410-08115}, we employ the F1 score between final answers and labels as evaluation metrics for information exchange tasks. For debate tasks, we utilize exact match accuracy (ARC-C, MMLU). For the deepsearch task, following~\citet{wu2025webwalker}, we utilize Qwen2.5-72B-Instruct~\citep{qwen2025qwen25technicalreport} to verify whether the answers were consistent with the correct answers.

\textbf{Baseline} For static scenarios, we compare our methods with: (1) Chain-of-Thought (CoT)~\citep{DBLP:conf/nips/Wei0SBIXCLZ22}: single agent pipeline which enables complex reasoning to derive the final answer.
(2) Multi-Agent Debate (MAD)~\citep{DBLP:conf/icml/Du00TM24}: multi-agent pipeline where different reasoning processes are discussed multiple rounds to arrive at the final answer. (3) AutoForm~\citep{DBLP:conf/emnlp/ChenYYSQYXL024}: multi-agent pipeline where the agents utilize non-nature language formats in communication to improve efficiency. For the dynamic scenario, following~\citet{Li2025WebThinker}, we evaluate several pipeline methods within this setting, including (1) Direct Reasoning, (2) RAG workflow and its variant~\citep{Li2025WebThinker}, and (3) Search-o1~\citep{DBLP:journals/corr/abs-2501-05366}. In both scenarios, we compare DITS with multi-agent optimization method Optima~\citep{DBLP:journals/corr/abs-2410-08115}: a framework that enhances communication efficiency and task effectiveness through Supervised Finetuning and Direct Preference Optimization. It has three variants, namely Optima-iSFT, Optima-iDPO, and Optima-iSFT-DPO. We follow the iSFT-DPO variant of Optima and improve its data synthesis and selection process to obtain DITS-iSFT-DPO.

\textbf{Implementation Details} We utilize the Llama-3-8B-Instruct~\citep{dubey2024llama} as the base model for static scenarios. Experimental results for other base models are provided in Appendix~\ref{appendix:base models}. For the dynamic scenario, we employ the QwQ-32B~\citep{QwenTeam2024Qwq} as the base model due to the task complexity.  The interaction ends when either a special token marks the final answer or the maximum number of turns is reached.
Unless otherwise specified, we set the hyperparameters to $\alpha=0.5$ and $\gamma=1$. When collecting influence scores via single-step gradient descent, we utilize LoRA (Low-Rank Adaptation)~\citep{DBLP:conf/iclr/HuSWALWWC22}. A validation set of size 20 is used in all experimental settings. We set the expansion time $d=3$ and repeat time $k=8$ for all datasets. More details are provided in the Appendix~\ref{appendix:training details}. 

%% file: table/main_table.tex
\begin{table*}[t]
    \centering
    \begin{tabular}{lcccccc}
    \toprule
    & \multicolumn{4}{c}{\textbf{Information Exchange}} & \multicolumn{2}{c}{\textbf{Debate}} \\
    \cmidrule(lr){2-5} \cmidrule(lr){6-7}
     \textbf{Method} & \multicolumn{1}{c}{\textbf{HotpotQA}} & \multicolumn{1}{c}{\textbf{2WMH QA}}  &\multicolumn{1}{c}{\textbf{TriviaQA}} & \multicolumn{1}{c}{\textbf{CBT}}& \multicolumn{1}{c}{\textbf{ARC-C}}&\multicolumn{1}{c}{\textbf{MMLU}} \\
    
    \midrule
    CoT & 25.6  &20.5  &59.8  &43.4 & 65.2  & 46.0 \\
    \midrule
    MAD &  28.4  &25.9& 71.0 & 53.8 & 71.4 & 51.5\\
    AutoForm & 28.2  &24.7  & 60.9  & 35.0  & 60.2  & 43.8 \\
    \midrule
    Optima-iSFT 
   
     & 54.5 & 72.4  & 71.9  & \underline{71.8} & 74.1  & 56.8\\
    Optima-iDPO 
     
     & 52.5  & 66.1  & 69.3  & 66.7 & 74.5  & 59.6 \\
    Optima-iSFT-DPO 
      & \underline{55.6}  & \underline{74.2}  & \underline{77.1}  & 70.1 & \underline{77.1}  & \underline{60.2} \\
      \midrule
    DITS-iSFT-DPO  & \textbf{57.2} & \textbf{76.0} & \textbf{78.4} & \textbf{72.0}  & \textbf{77.6} & \textbf{60.5} \\

    \bottomrule
    \end{tabular}
    \caption{\textbf{Performance comparison across Information Exchange and Debate tasks.} Best results are indicated in \textbf{bold}, and second-best results are \underline{underlined}. The baseline results are taken from~\citet{DBLP:journals/corr/abs-2410-08115}.}
    \label{tab:main-table}
    
\end{table*}

%% file: section/6_Experiments.tex
\section{Evaluation Results}
\label{section:experiments}
In this section, we first evaluate the effectiveness of DITS (§~\ref{subsection:main_results}). Then we demonstrate the superiority of data influence through ablation studies (§~\ref{subsection:single_iteration}) and explore the impact of synthesis scaling on data quality (§~\ref{subsection:synthesis time scaling}). Finally, we analyze the effects of selection ratio, iteration times, and validation set size (§~\ref{subsection:selection_ratio}).

\subsection{Overall Performance}
\label{subsection:main_results}

\textbf{The Static Scenarios} In Table~\ref{tab:main-table}, we compare our method DITS-iSFT-DPO with the baseline approaches on both the Information Exchange and Debate tasks. Across all datasets, our method achieves consistent improvement over the baselines, demonstrating the effectiveness and generalizability of DITS. Compared to the single agent CoT approach, our method delivers an average performance enhancement of 91\%. In the Information Exchange task, our method outperforms the advanced multi-agent approach Optima-iSFT-DPO by an average margin of 2.5\%.

\input{table/webwalker}
\textbf{The Dynamic Scenario}
\label{subsection:dynamic_results}
In dynamic scenarios, we adopt the WebThinker framework~\citep{Li2025WebThinker} to structure the process into a collaborative system comprising three agents: a task analysis agent, a search intent generation agent, and a web content analysis agent. This framework empowers the agents to autonomously conduct web searches, deeply analyze web content, and dynamically adjust their collaboration strategies. For search, we use the Serper API\footnote{https://serper.dev/}, retrieving the top 10 search results (k=10). In Table~\ref{tab:webwalker}, we observe that the Webthinker framework for dynamic multi-agent collaboration outperforms traditional single-agent approaches and simple RAG methods. Furthermore, fine-tuning multiple agents within the collaborative framework enhances coordination efficiency. Notably, the DITS method surpasses all baseline models, highlighting its effectiveness and robustness.

\input{table/single_iteration}

\begin{figure*}
    \centering
    \includegraphics[width=0.19\linewidth]{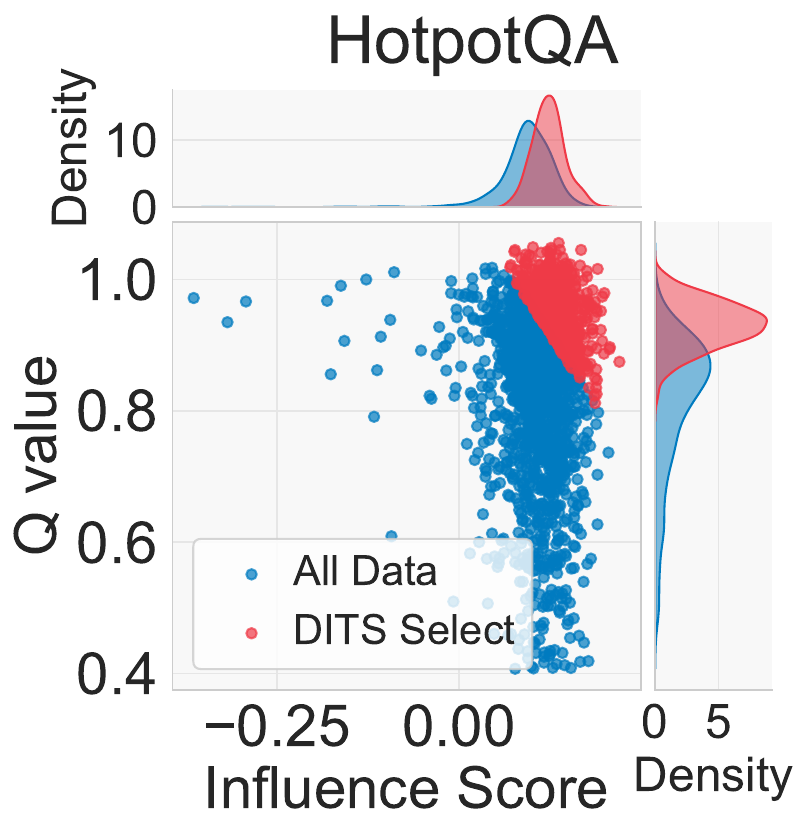}
    \includegraphics[width=0.19\linewidth]{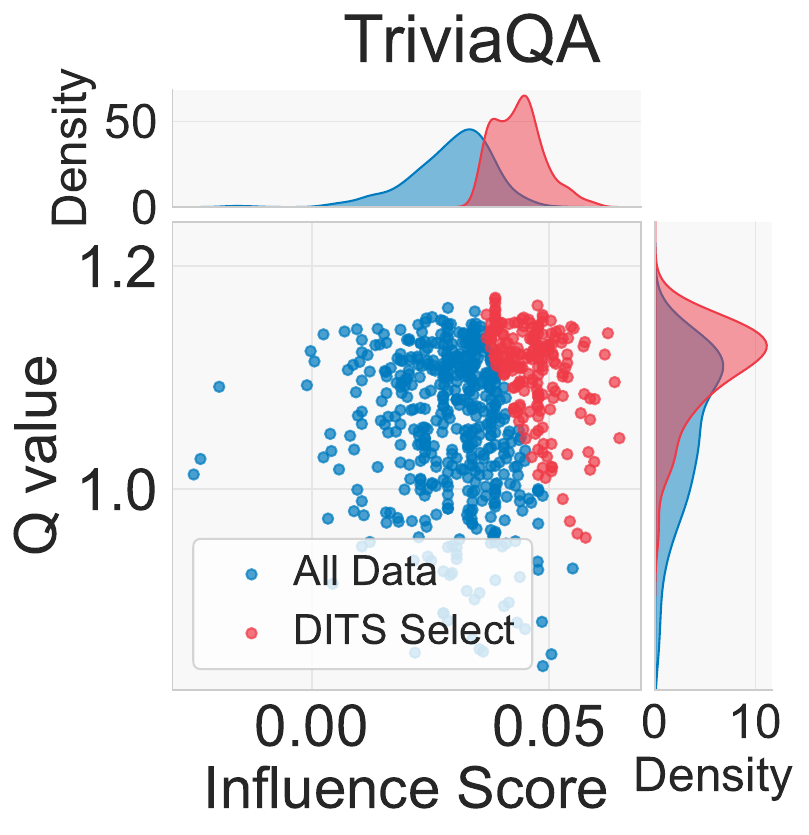}
    \includegraphics[width=0.19\linewidth]{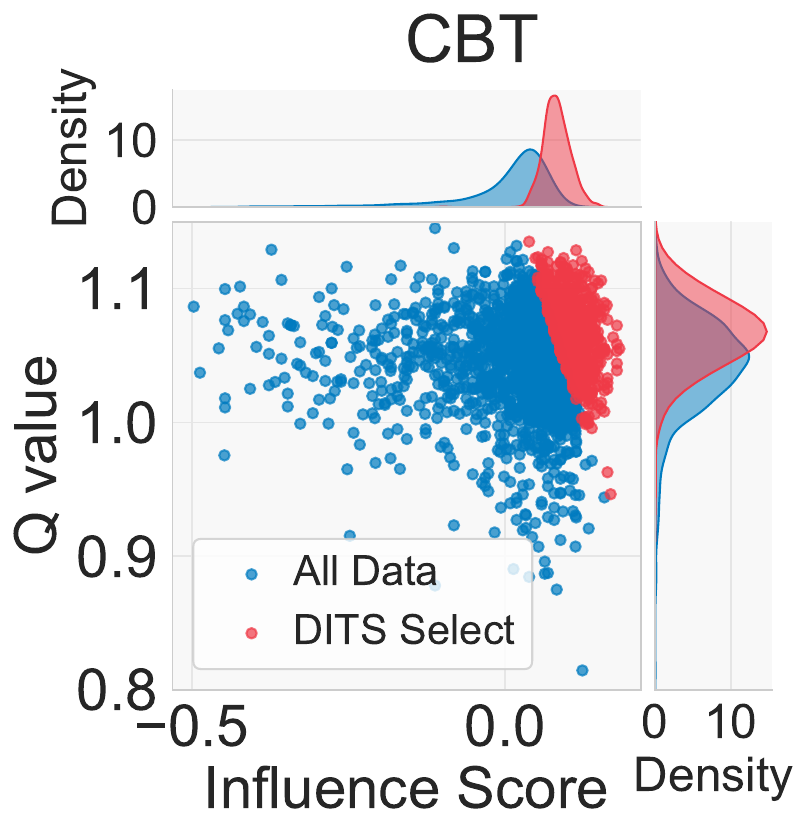}
    \includegraphics[width=0.19\linewidth]{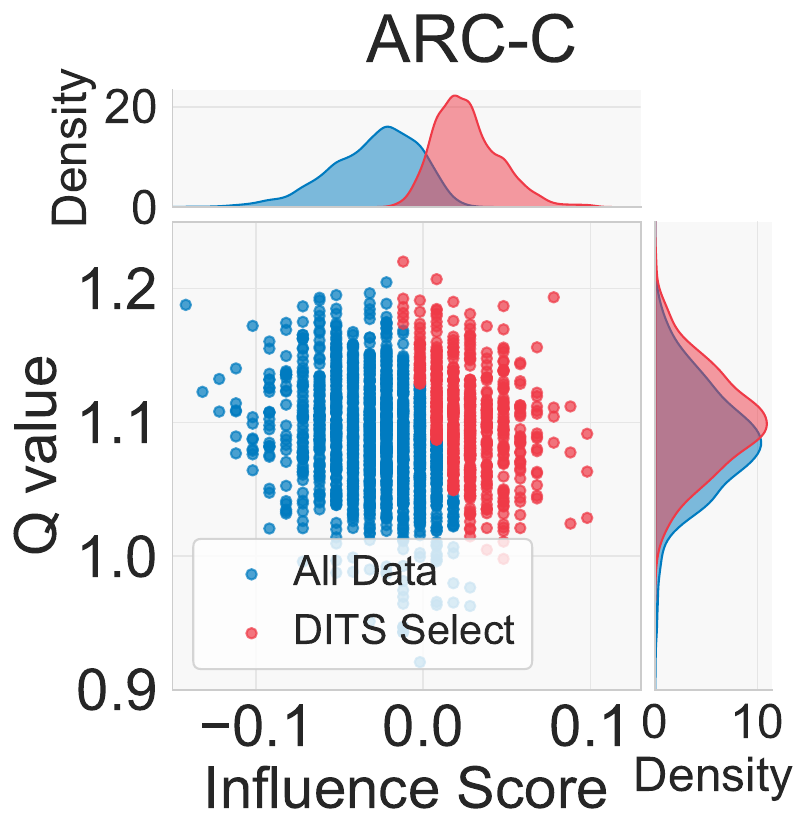}
    \includegraphics[width=0.19\linewidth]{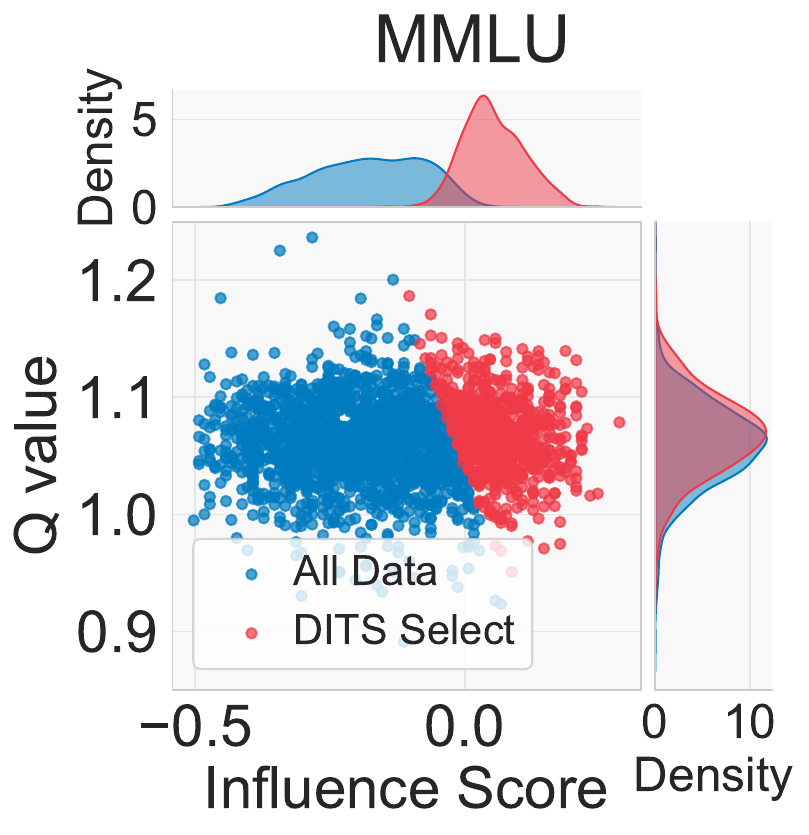}
    \caption{The scatter plot and density plots of Q-values and influence scores for synthetic data. The top 30\% of the data selected by DITS is highlighted in red.}
    \label{fig:distribution_analysis_appendix}
\end{figure*}

\subsection{Influence Function Analysis}
\label{subsection:single_iteration}

To provide a detailed comparison of the effectiveness of the influence function, we present the results of different data selection methods in Table~\ref{tab:single-iteration}. The experiments are conducted in a single iteration. The Base method represents the multi-agent framework performance with the base model Llama-3-8B-Instruct. The Optima-DPO and Optima-RPO methods utilize the dataset $\mathcal{D}_{\text{tr}}$ sampled through the MCTS approach in Optima to train the model using DPO loss~\citep{DBLP:conf/nips/RafailovSMMEF23} and RPO loss~\citep{DBLP:journals/corr/abs-2404-19733}, respectively. Random Select refers to training on the data randomly sampled from $\mathcal{D}_{\text{tr}}$ with DPO loss, while Q-value Select involves selecting the top-ranked data based on Q-values for training. DITS employs the influence score in Eq.~\eqref{equation:hybrid score} to select the top-ranked data for training, where the variant $\gamma=1$ integrates both Q-value and influence score, and the variant $\gamma=0$ relies solely on the influence score for data selection. For a fair comparison, we set the selection ratio as 50\% for all methods.

\textbf{Ablation Study} As shown in Table~\ref{tab:single-iteration}, we observe that (1) The DITS method achieves consistent performance improvements across all datasets compared to using the full dataset, indicating that the original MCTS-generated dataset contains noisy and lower-quality data. This suggests that further enhancing data quality is beneficial for model performance.  (2) Selecting data based on influence scores outperforms both random selection and Q-value-based selection, highlighting its superior effectiveness in enhancing data quality. To further explore the underlying reasons for this improvement, the following paragraph provides an in-depth analysis of the data distribution. (3) For the Information Exchange task, the variant $\gamma=0$ achieves the best performance, while the variant $\gamma=1$ achieves suboptimal results. In contrast, on the Debate task, the variant $\gamma=1$ generally performs the best. This discrepancy is attributed to the fact that the evaluation metric for the Information Exchange task is F1-score, which introduces more noise into the estimated Q-values, resulting in lower quality in selecting data.

\input{table/time_comparson}

\begin{figure*}
    \centering
    \subfigure[]{\includegraphics[width=0.50\textwidth]{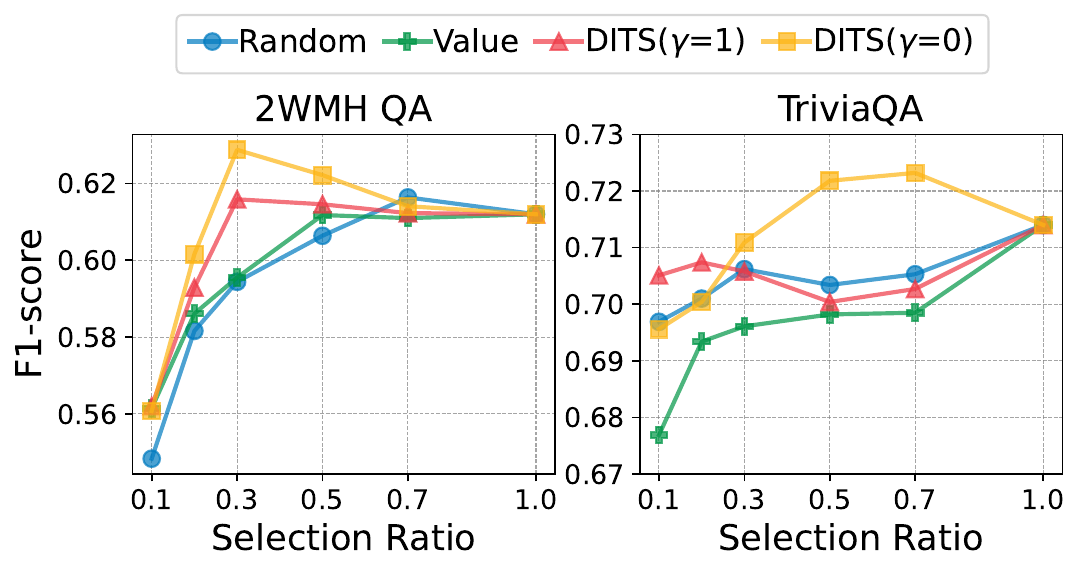}}
    \subfigure[]{\includegraphics[width=0.24\textwidth]{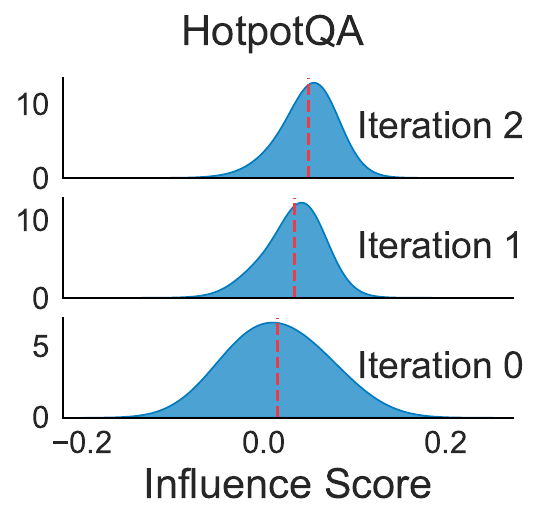}\includegraphics[width=0.24\textwidth]{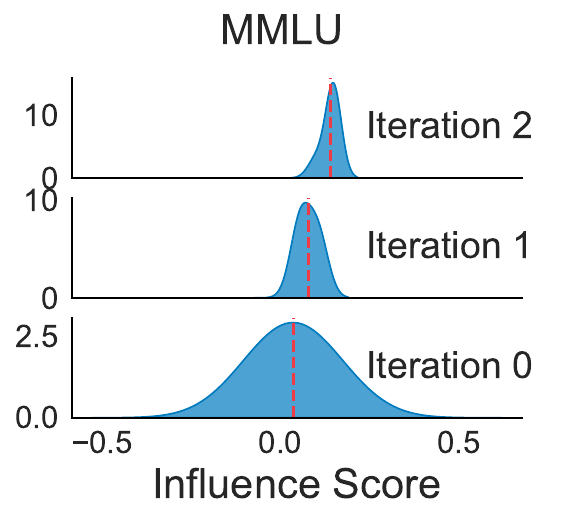}}
    \caption{(a) The effect of hyperparameter selection ratio $\alpha$ of DITS on the 2WMH QA and TriviaQA datasets. (b) The distribution of synthetic data influence scores across different iterations on the HotpotQA and MMLU datasets, with the mean of the distribution highlighted by a red dashed line.}
    \label{fig:selection_ratio}
    
\end{figure*}

\textbf{Distribution Analysis} 
To provide an in-depth analysis of the advantages of using the influence score for data selection, we visualize the distributions of Q-values and influence scores in Figure~\ref{fig:scaling} (a) and Figure~\ref{fig:distribution_analysis_appendix}, highlighting the distribution of the top 30\% data points selected by our methods with $\gamma=1$. From the figures, we observe that: (1) There are discrepancies between the influence score and Q-value, which reveals that the Q-value is not perfectly aligned with training needs. This highlights the importance of integrating the influence score into the MCTS process and data selection process. (2) The data selected by our methods exhibit high influence scores and Q-values, indicating that DITS is capable of selecting high-quality data.

\subsection{Synthesis Time Scaling}
\label{subsection:synthesis time scaling}

In this study, we empirically demonstrate that increasing the synthesis budget during the data synthesis phase enhances model performance, as shown in Figure~\ref{fig:scaling} (d) and Section~\ref{subsection:synthesis time scaling}. Specifically, the figure highlights three key observations: (1) Allocating a larger synthesis budget, which extends rollout times and increases the number of expansions, will generate more high-quality data, thereby improving model performance. (2) We validate that allocating resources to influence score estimation can indeed lead to better performance improvements. This is attributed to the fact that the influence score is more aligned with training needs. This underscores the capability of our method to enhance the efficiency of synthesizing training data within a vast action space.
(3) The performance gains from a sixteenfold increase in the synthesis budget are notably smaller compared to the improvements achieved through three times iterative data synthesis and training, as detailed in Table~\ref{tab:main-table}. This comparison highlights the efficiency and effectiveness of the iterative approach. 



\textbf{Efficiency Analysis} We provide an empirical computation cost comparison between DITS and Optima. Using the 2WMH QA dataset as an illustrative example, we compare the training costs per iteration across different settings. As shown in Table~\ref{table:time_comparson},  although the incorporation of data influence estimation introduces additional computational overhead, we argue that this extra cost is justified by its greater effectiveness in improving model performance. A further efficiency comparison with conventional data influence estimation methods is provided in Appendix~\ref{appendix:efficiency analysis}.


\subsection{Hyperparameter Analysis}
\label{subsection:selection_ratio}

\textbf{Selection Ratio} We first investigate the impact of the selection ratio hyperparameter $\gamma$ on model performance. We conduct experiments on two Information Exchange tasks: 2WMH QA and Trivia QA datasets, and present the results in Figure~\ref{fig:selection_ratio}. We compare Optima-DPO (random Select and Q-value Select) with DITS ($\gamma=0$) and DITS ($\gamma=1$). From the figure, we observe that: (1) Across different selection ratios, DITS consistently outperforms Optima-DPO, demonstrating that our method can select data more beneficial for model training and exhibits strong generalization ability. (2) When an appropriate selection ratio is chosen, the performance of DITS surpasses that of using the full dataset, indicating the presence of noise in the original MCTS-generated data and the potential for further improving data quality. (3) When the selection ratio is very small, the performance of all methods declines, indicating that training set size is also crucial for achieving optimal performance. This suggests that an overly small yet high-quality dataset may not be sufficient to train a good model.







\textbf{Iteration Times} To gain deeper insights into the iterative data synthesis and training process, we analyzed the distribution of influence scores for synthetic data across different iterations on the HotpotQA and MMLU datasets, as shown in Figure~\ref{fig:selection_ratio} (b). The mean of each distribution is highlighted. From the figure, we observe the following trends: (1) As the number of iterations increases, the mean influence score gradually rises, indicating an improvement in the quality of synthetic data. This suggests that the iterative process enhances data quality by refining the model, creating a positive feedback loop that makes data synthesis more effective. (2) With more iterations, the distribution of influence scores becomes more concentrated, suggesting that the model trained on synthetic data achieves more stable quality on specialized tasks. However, this may come at the cost of reduced data diversity.

We further analyze model performance over training iterations, the impact of validation set size, and compare data selection strategies based on influence scores. Details are provided in Appendix~\ref{appendix: ablation study}.







\begin{figure}
    \centering
    \includegraphics[width=0.85\linewidth]{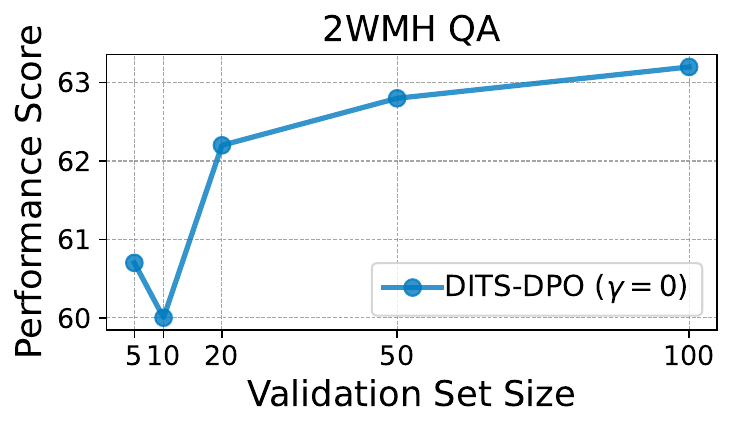}
    
    \caption{The effect of hyperparameter validation size $V$ on the performance of DITS.}
   
    \label{fig:validation_size}
\end{figure}

\textbf{Validation Set Size} To investigate the relationship between DITS and validation set size, we conducted additional experiments on 2WMH QA. As shown in the Figure~\ref{fig:validation_size}, the performance of DITS-DPO improves with an increasing validation set size, indicating that a larger validation set provides a more accurate estimation. However, since a larger validation set demands a higher synthesis budget, a trade-off is necessary in practical applications.

%% file: table/webwalker.tex
    

\begin{table}[t]
    \centering
    \begin{tabular}{lc}
    \toprule
       {Models} & WebWalker \\
    \midrule
    Direct Reasoning & 4.3 \\
    RAG Workflow & 31.2\\
    $\quad$- w/ Query Planning & 32.5 \\
    $\quad$- w/ Iterative RAG & 31.5 \\
    Search-o1 & 34.1\\
    \midrule
    WebThinker  & \\
    $\quad$- Base & 37.0 \\
    $\quad$- Optima-SFT & 46.0 \\
    $\quad$- Optima-DPO & 46.6 \\
    $\quad$- DITS-DPO & \textbf{47.2} \\
    \bottomrule
    \end{tabular}
    \captionof{table}{Performance comparison on DeepSearch task. Best results are indicated in \textbf{bold}. }
    \label{tab:webwalker}
   \vspace{-10pt}
\end{table}

%% file: table/single_iteration.tex
\begin{table*}[t]
    \centering
    \begin{tabular}{lcccccc}
    \toprule
    & \multicolumn{4}{c}{\textbf{Information Exchange}} & \multicolumn{2}{c}{\textbf{Debate}} \\
    \cmidrule(lr){2-5} \cmidrule(lr){6-7}
     \textbf{Method} & \multicolumn{1}{c}{\textbf{HotpotQA}} & \multicolumn{1}{c}{\textbf{2WMH QA}}  &\multicolumn{1}{c}{\textbf{TriviaQA}} & \multicolumn{1}{c}{\textbf{CBT}}  & \multicolumn{1}{c}{\textbf{ARC-C}}&\multicolumn{1}{c}{\textbf{MMLU}} \\
    
    \midrule
    Base & 28.2  & 24.7  & 60.9  & 35.0 & 60.2 & 43.8 \\
    Optima-SFT  & 45.2  & 59.7   & 68.8  & 50.7 & 68.2 & 50.3 \\
    \midrule
    Optima-RPO & 50.4 & 60.6 &68.4 &\underline{59.1}  & 72.2  & \underline{52.1} \\
     
    Optima-DPO   & 46.6 & 61.2 &70.9  & 57.2 & 71.5 & 51.6 \\
    $\quad$- Random Select            &51.5 & 60.6 & 70.3  & 58.0  &74.0 &51.1\\
    
    $\quad$- Q-value Select          &50.5  & 61.1 &69.8  &58.6 &73.7 &50.2 \\
    \midrule
    DITS-DPO  && &&  && \\
    $\quad$- $\gamma=0$ &\textbf{53.1}  & \textbf{62.2} &\textbf{72.2}  &\textbf{59.6}  &\underline{74.2} &50.8 \\
    $\quad$- $\gamma=1$    &\underline{52.8}  & \underline{61.5}  & \underline{71.0} &\underline{59.1}  &\textbf{74.5}  &\textbf{52.3} \\
    \bottomrule
    \end{tabular}
    \caption{\textbf{Single iteration performances across Information exchange and Debate tasks.} Best results are indicated in \textbf{bold}, and second-best results are \underline{underlined}.}
    \label{tab:single-iteration}
\end{table*}

%% file: table/time_comparson.tex
\begin{table*}[t]
\centering
\resizebox{\linewidth}{!}{%
\begin{tabular}{lcccccc}
\toprule
Method & \makecell{Synthesis Budget\\ (Token)} & \makecell{\# Samples} & \makecell{Synthesis Cost\\ (GPU Hours)} & \makecell{Training Cost\\ (GPU Hours)} & \makecell{Total Cost\\ (GPU Hours)} & \makecell{Performance\\ (F1 score)} \\
\midrule
Optima-DPO & $1.67 \times 10^7$ & 17000 & 82 & 16 & 98 & 0.607 \\
\midrule
Optima-DPO & $3.34 \times 10^7$ & 34000 & 165 & 30 & 195 & 0.610 \\
\midrule
DITS-DPO & $2.00 \times 10^7$ & 8500 & 98 & 8 & 106 & 0.612 \\
\bottomrule
\end{tabular}
}
\caption{Comparison of training costs and performance between DITS and Optima on the 2WMH QA dataset.}
\label{table:time_comparson}
\end{table*}

%% file: section/7_Conclusion.tex
\section{Conclusion}
\label{section:conclusion}

In this work, we propose DITS, a novel multi-agent data self-training framework that integrates influence scores into MCTS to guide tree search and data selection. By leveraging influence scores and proposing a novel estimation method, we effectively identify the most impactful data for system improvement, thereby enhancing model performance. Meanwhile, we derive an efficient influence score estimation method for non-differentiable metrics through gradient-to-inference conversion. This approach substantially reduces computational overhead through inference computation and allows us to estimate influence scores to achieve a more efficient data synthesis process. 
Our approach introduces new perspectives and scaling dimensions for data synthesis, highlighting the impact of training data on performance rather than its correctness.


%% file: section/8_Limitation.tex
\section{Limitation}
\label{section:limitation}
Our method demonstrates strong performance in both static and dynamically coordinated multi-agent settings. However, the current study does not cover more open-ended collaboration regimes, such as dynamic agent spawning or emergent team formation. These settings can induce substantially greater variance in agent interactions and coordination patterns, making controlled and reproducible evaluation considerably more challenging. In addition, DITS is primarily designed for offline, training-time data selection, where the overhead of multiple inference passes is an acceptable trade-off for improved data quality. Under strict latency constraints, however, the current influence estimator may be too costly for real-time or streaming data quality assessment. Developing evaluation protocols for open-ended collaboration, as well as lightweight single-pass approximations or end-to-end trainable influence models for more efficient quality estimation, remains an important direction for future work.

%% file: section/Appendix.tex
\input{section/2_RelatedWork}

\section{Theoretical Analysis}
\label{appendix: theoretical analysis}

In this section, we illustrate the relationship between influence scores and model performance, where the selection of the most influential data points maximizes the model's validation performance. Concretely, we first extend the definition of influence score to a data group $U\subset \mathcal{D}_{tr}$ as:
$$ \theta_{\epsilon,U}^*= \underset{\theta}{\arg\min} \sum_{z\in \mathcal{D}_{\text{tr}}} \frac{1}{|\mathcal{D}_{\text{tr}}|} \mathcal{L}_{\text{tr}}(z, \theta) + \epsilon \sum_{z\in U}\mathcal{L}_{\text{tr}}(z, \theta). $$

Following~\cite{DBLP:conf/icml/KohL17}, under the first-order approximation, we have 
$$\mathcal{I}_{\mathcal{L}_{\text{tr}}}(U, \mathcal{D}_{\text{tr}}) \stackrel{\text{def}}{=} \frac{d \mathcal{L}_{\text{tr}}(U, \theta_{\epsilon, U}^*)}{d\epsilon} \bigg|_{\epsilon=0} \approx \sum_{z\in U} \mathcal{I}_{\mathcal{L}_{\text{tr}}}(z, \mathcal{D}_{\text{tr}}),$$

where the influence score of a group of data points can be represented as the sum of the influence score of individual data points. For DITS, we adopt a similar approximation:
$$ \mathcal{I}_{\mathcal{F}_{\text{val}}}(U, \mathcal{F}_{\text{val}}) \approx \sum_{z\in U} \mathcal{I}_{\mathcal{F}_{\text{val}}}(z, \mathcal{D}_{\text{val}}).$$
Thus, the selection of the most influential data points maximizes validation performance.

\section{Algorithm}
\begin{algorithm}[t]
\caption{DITS-iSFT-DPO}
\label{alg:DITS-isft-dpo}
\begin{algorithmic}[1]
\Require Initial model $\theta_\text{init}$, problem Set $\mathcal{D}$, validation Set $\mathcal{D}_{\text{val}}$, and max iterations $T$
\Ensure parameter $\theta_T$
\State $\theta_0 \gets \theta_\text{init}$
\For{$t = 1$ to $T$}
    \State $D_t^{SFT}$ $\gets$ SFTDataCollect($\theta_{t-1}$) \Comment{Following \cite{DBLP:journals/corr/abs-2410-08115}}
    \State $\theta_t$ $\gets$ SFT($D_t^{SFT}$, $\theta_\text{init}$) \Comment{Following \cite{DBLP:journals/corr/abs-2410-08115}}
    \State $\mathcal{D}_t^\text{DPO} \gets \emptyset$
    \ForAll{$p_i \in \mathcal{D}$}
        \State $\mathcal{D}_i^\text{DPO} \gets \text{MCTSSynthesis}(\theta_t, p_i)$ 
        \State $\mathcal{I}_{\mathcal{F}_{\text{val}}}\gets \text{DataInfluenceCollect}(\mathcal{D}_{\text{val}})$ 
        \State $\mathcal{D}_t^\text{DPO} \gets \mathcal{D}_t^\text{DPO} \cup \mathcal{D}_i^\text{DPO}$
    \EndFor
    \State $\mathcal{D}_t^{DPO} \gets \text{InfluSelection}(\mathcal{D}_t^{DPO}, \mathcal{I}_{\mathcal{F}_{\text{val}}}$)
    \State $\theta_{t}$ $\gets$ DPO($\mathcal{D}_t^{DPO}, \theta_t)$
\EndFor
\State \textbf{return} $\theta_T$
\end{algorithmic}
\end{algorithm}

The DITS-iSFT-DPO algorithm (Algorithm~\ref{alg:DITS-isft-dpo}) iteratively refines a model by alternating between SFT and DPO. In each iteration, the model is fine-tuned using new data collected via SFT. Then, DPO training data is generated through MCTS synthesis and filtered using influence scores from a validation set. The model is updated with DPO to better align with preferences, leading to progressive improvement.

\section{Method Details}
\label{appendix:method_details}

\subsection{Reward Function}
Following Optima~\citep{DBLP:journals/corr/abs-2410-08115}, we define each trajectory $\tau_i$ is then evaluated using a reward function $R: \mathcal{T} \rightarrow \mathbb{R}$:
\begin{equation}
    R(\tau_i^j) = R_\text{task}(\tau_i^j) - \lambda_\text{token} R_\text{token}(\tau_i^j) + \lambda_\text{loss} \frac{1}{R_\text{loss}(\tau_i^j)}.
    \label{eq:reward}
\end{equation}
Here, $R_\text{task}: \mathcal{T} \rightarrow \mathbb{R}$ is the task-specific performance metric, $R_\text{token}(\tau_i^j) = \frac{\#\text{Tokens}(\tau_i^j)}{\max_k(\{\#\text{Tokens}(\tau_i^k)\}_k)}$ is the normalized token count, and $R_\text{loss}(\tau_i^j) = g\big(\mathcal{L}(\mathcal{M}_\text{base}, d_i, \tau_i^j)\big)$ is based on the language modeling loss of the base model $\mathcal{M}_\text{base}$. The positive coefficients $\lambda_\text{token}$ and $\lambda_\text{loss}$ are hyper-parameters. More details can refer to Optima~\citep{DBLP:journals/corr/abs-2410-08115}.

\subsection{Initial Data Filtering}
For the preference data pairs obtained from the MCTS tree, we follow the Optima~\citep{DBLP:journals/corr/abs-2410-08115} by initially filtering the data pair $(s, a_i^h, a_i^l)$. Specifically, we select pairs that satisfy: (1) $R(s, a_i^h) > \lambda_{\text{dpo-filter}}$. (2) $R(s, a_i^h)- R(s, a_i^l)>\lambda_{\text{dpo-diff}}$. (3) For preference pairs starting with the same problem $p$, we rank these pairs based on their Q-values and select the top 50\% of the pairs.

\input{table/efficiency_comparison}
\input{table/selection_strategy}

\section{Dataset Details}\label{appendix:dataset}

In the information exchange setting, the relevant context is divided between two agents. The agents must identify the relevant information and communicate with each other to derive the final answer. This setting includes HotpotQA~\citep{DBLP:conf/emnlp/Yang0ZBCSM18}, 2WikiMultiHopQA (2WMH QA)~\citep{DBLP:conf/coling/HoNSA20}, TriviaQA~\citep{DBLP:conf/acl/JoshiCWZ17}, and CBT~\citep{DBLP:journals/corr/HillBCW15}. In the debate setting, two agents work together to solve a task: one agent proposes solutions, while the other evaluates their correctness. 
The debate setting includes ARC's challenge set (ARC-C)~\citep{DBLP:journals/corr/abs-2102-03315} and MMLU~\citep{DBLP:conf/iclr/HendrycksBBZMSS21}. Unlike static scenarios, where multi-agent collaboration follows a predetermined sequence, the dynamic DeepSearch setting features agents that autonomously determine and continuously adjust their collaboration strategies based on the evolving task, enabling truly adaptive and intelligent teamwork. The DeepSearch task involves collaboration among a task analysis agent, a search intent generation agent, and a web content analysis agent. The DeepSearch setting includes WebWalker~\citep{wu2025webwalker}. We use 0-shot for all benchmarks.



\section{Efficiency Analysis}
\label{appendix:efficiency analysis}

In this section, we quantify the computational cost of the different influence score estimation methods. For the forward pass of LLM, the computational cost is $2NS+4LHS^2$, where $S$ is the sequence length, $N$ is the number of model parameters, $L$ is the number of model layers, and $H$ is the embedding dimension of the model. For small $S$ (e.g., 2000), the second term is negligible, making the cost per token $2N$. The backward pass doubles this cost to $4N$. During inference with KV cache, generating one token also costs $2N$.

We compare two classic gradient-based data influence methods, TRAK~\citep{DBLP:conf/icml/ParkGILM23} and LESS~\citep{DBLP:conf/icml/XiaMGA024}, under the following assumptions: average sequence length $S=2000$,  validation set size $V=20$, model parameters $N=8B$, projection dimension $d=8192$ for TRAK and LESS, and $R=4$ checkpoints for LESS.

The computational costs (in FLOPs) for estimating the influence score of one data point are described in Table~\ref{tab:efficiency}. As shown in the table, our method exhibits superior efficiency compared to traditional approaches. This advantage primarily stems from the fact that gradient-based methods require gradient computation on validation data and necessitate parameter projection onto a low-dimensional subspace.

\section{Ablation Study}
\label{appendix: ablation study}

\begin{figure}[t]
 \centering
\includegraphics[width=0.48\textwidth]{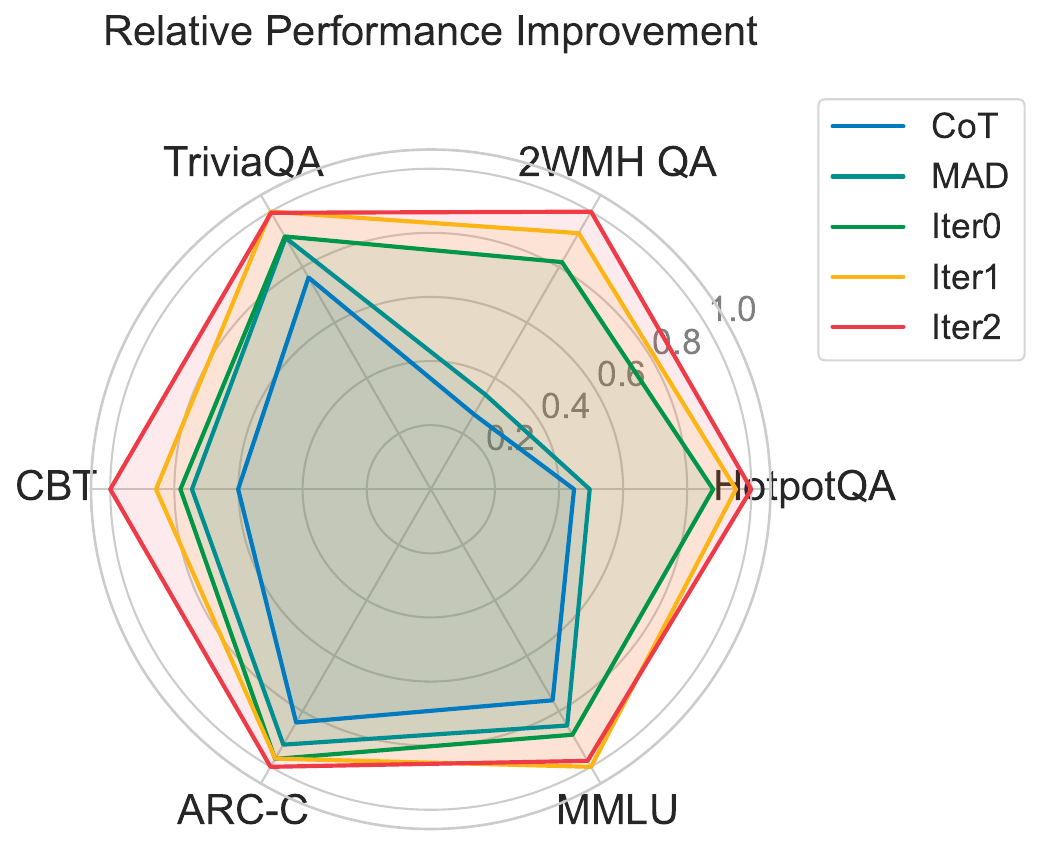}
    \caption{The relative performance improvement of DITS-iSFT-DPO across all datasets at different iterations. The best performance of each dataset is set as 1.0.
    }
    \label{fig:iteration}
\end{figure}

\subsection{Iterative Times Analysis} Using the performance of CoT as the baseline, we report the average relative performance improvement of our method, DITS-iSFT-DPO, across all datasets per iteration and present the results in Figure~\ref{fig:iteration}. From the figure, we observe that: (1) Our method achieves an average improvement of 91\% compared to the single-agent CoT approach and an improvement of 64\% over the multi-agent MAD method, demonstrating the effectiveness of our approach. (2) As the number of iterations increases, the average performance continues to improve. Since we start training from the same initial model in each iteration, this indicates that training better models and subsequently synthesizing data can consistently enhance the quality of the generated data and improve the final performance.

\input{table/robustness_table}

\input{table/base_model}

\subsection{Selection Strategy Analysis} 

The influence scores are estimated on the validation set according to Eq~\eqref{eq:influence_estimation}. A larger size of validation set V yields a more accurate estimation but requires a higher synthesis budget.

Data points with high influence scores but low Q-values can still contribute to model training, and we have provided an example in the Appendix~\ref{appendix:case study}. Empirical results, as shown in the Table~\ref{tab:selection_strategy} (V= 20), further validate this conclusion. However, when the estimated influence scores contain noise (due to a small V), supplementing the selection with high Q-values can help filter higher-quality data.

Data points with high Q-values but low influence scores represent data that the model already handles well. Further training on them risks overfitting and hinders model advancement. Empirically, these data points significantly degrade model performance, performing even worse than random select.

Overall, the contribution of data points to model training fundamentally depends on the influence scores. The correlation coefficient between Q-values and influence scores is approximately 0.1, indicating that traditional methods relying solely on high Q-values are insufficient for selecting optimal training data. When constrained by a limited synthesis budget—leading to less accurate influence score estimation—leveraging high Q-values as an auxiliary filtering criterion can improve data selection.

\subsection{Robustness Analysis}

Since the influence score is estimated based on a validation set, its accuracy may depend on the choice and size of the validation data. As shown in Section~\ref{subsection:selection_ratio}, the performance of DITS-DPO generally improves with a larger validation set, suggesting that more validation examples can provide a more accurate influence estimate. However, increasing the validation set size also requires a higher synthesis budget, leading to a practical trade-off between estimation quality and computational cost.

To further examine robustness, we conduct two additional analyses. First, we randomly sample two validation sets of different sizes, containing 20 and 200 examples, respectively, and compute the influence score of each data point based on these two sets. The Pearson correlation coefficient between the resulting scores is 0.912, indicating that even a small randomly selected validation set already provides a highly consistent signal for influence estimation. Second, on the 2WMH QA dataset, we construct five different validation sets of the same size (20 examples), compute influence scores separately, and train DITS-DPO based on the resulting data selection. As shown in Table~\ref{tab:validation_robustness}, the small variation in performance demonstrates that DITS-DPO is robust to the specific choice of validation data. Overall, these results suggest that the proposed influence-based data selection method remains stable under different validation sets, while larger validation sets can further improve estimation reliability.

\input{table/hyper_parameter_main}
\input{table/hyper_parameter_single}

\subsection{Results with Different Base Models}
\label{appendix:base models}
In this section, we evaluate DITS with Llama-3.1-8B-Instruct~\citep{dubey2024llama} and Qwen2.5-7B-Instruct~\citep{qwen2025qwen25technicalreport} across information exchange and debate tasks. As shown in Table~\ref{tab:base-model}, we observe that across most datasets, DITS outperforms all baseline methods, demonstrating strong generalization and robustness. Moreover, experiments show that incorporating more powerful base models and applying task-specific fine-tuning further enhances performance, suggesting that stronger base models promote more effective multi-agent cooperation.

\section{Training Details}
\label{appendix:training details}

 The hyperparameters we used are shown in Table~\ref{tab:hyper_parameters_main} and Table~\ref{tab:hyper_parameters_single}.

\section{Case Study}
\label{appendix:case study}
\input{table/case}
As illustrated in Figure~\ref{tab:cases}, we present a comparative case study highlighting the differences in data selection outcomes when using Q-value versus influence score for the same task. The task—"Which film has the director who was born later, Eyes of the Forest or Stardust on the Sage?"—was analyzed through agent dialogues between Alice and Bob.  

In the Q-value-selected data pair, the dialogue history efficiently conveyed the directors' birth dates within a single interaction round. The chosen response directly identified "Stardust on the Sage" as the correct answer using special token markers in the response, achieving an exceptionally high Q-value. Meanwhile, the rejected response, although redundant in restating first-round information, contained no errors, thereby maintaining a high Q-value. However, the minimal difference between the paired responses resulted in low influence scores, limiting their utility for model improvement.  

In contrast, the influence-score-selected data pair exhibited incomplete information sharing in the dialogue history. The chosen response correctly ruled out Hillyer as the director of "Stardust on the Sage" but required more information to get to the correct answer, leading to lower Q-values. More critically, the rejected response contained hallucinatory content—an outright factual error falsely attributing Katedza as the film’s director—which fundamentally obstructed correct reasoning and resulted in an extremely low Q-value. This high-contrast pair data holds significant pedagogical value, as it juxtaposes valid reasoning with critical hallucinations, thereby achieving superior influence scores.  

Our analysis reveals that while high-Q-value pairs ensure accurate answers, they may have low influence scores and contribute little to multi-agent training. Conversely, data pairs with pronounced contrasts in reasoning validity—despite both exhibiting lower Q-values—substantially enhance model robustness against hallucinations by explicitly demarcating errors. These findings strongly advocate prioritizing influence score metrics over Q-value in both data synthesis and tree search to maximize model performance.

\input{section/Impact}

%% file: section/2_RelatedWork.tex
\section{Related Work}
\label{section:relatedwork}

\subsection{LLM based MAS}
\label{subsection:LLM based MAS}
LLM-based multi-agent systems (MAS) have demonstrated remarkable capabilities in addressing complex problems in various tasks~\citep{DBLP:conf/iclr/HongZCZCWZWYLZR24, DBLP:conf/acl/IslamAP24, tran2025multiagentcollaborationmechanismssurvey}. These systems employ various collaborative strategies, including multi-agent debate~\citep{DBLP:conf/icml/Du00TM24, DBLP:conf/emnlp/Liang0JW00Y0T24} and role-based division of labor~\citep{DBLP:journals/corr/abs-2405-04219, DBLP:conf/naacl/WangMW0WJ24}. 
Researchers have explored several key approaches to improve the performance of multi-agent systems. One strategy focuses on expanding the diversity and scale of agents~\citep{DBLP:journals/corr/abs-2402-05120, DBLP:conf/acl/QianLLCDL0CSCXL24, DBLP:journals/corr/abs-2406-04692}, optimizing performance from a network architecture perspective. Another approach emphasizes enhancing prompt quality, such as refining system memory in frameworks like AutoGen~\citep{DBLP:journals/corr/abs-2308-08155} and BiLLP~\citep{DBLP:conf/sigir/Shi0ZGLZWF24} or improving instruction design and few-shot examples in Dspy~\citep{DBLP:journals/corr/abs-2310-03714, DBLP:conf/emnlp/Opsahl-OngRPBPZ24}. A third approach involves fine-tuning the parameters of the large models within the agents, which is the most effective yet challenging method. Optima~\citep{DBLP:journals/corr/abs-2410-08115} and MALT~\citep{DBLP:journals/corr/abs-2412-01928} have taken the first step in this direction by constructing preference training data pairs through estimating Q-values. MALT can be viewed as a special case of Optima.

\subsection{Monte Carlo Tree Search}
\label{subsection:MCTS}

MCTS is an advanced search algorithm capable of effectively balancing exploration and exploitation in decision-making processes. It gained significant attention following its success in AlphaGo~\citep{DBLP:journals/nature/SilverHMGSDSAPL16}. Subsequently, researchers have introduced MCTS into LLM reasoning tasks~\citep{DBLP:conf/emnlp/HaoGMHWWH23}, giving rise to two primary methodologies. The first approaches employ MCTS during the inference phase, prioritizing actions with the highest potential to yield correct outcomes~\citep{DBLP:journals/corr/abs-2408-03314, wu2024inferencescalinglawsempirical}. The second approaches leverage MCTS during the training phase to synthesize high-quality training data, with the goal of identifying data that maximizes the improvement in model performance~\citep{DBLP:journals/corr/abs-2408-06195, DBLP:journals/corr/abs-2405-00451, DBLP:journals/corr/abs-2406-07394, DBLP:journals/corr/abs-2406-03816}. 
These approaches mainly rely on estimated Q-values to guide the exploration of the synthesis data space. 

\subsection{Influence Function}
\label{subsection:Influence function}

The influence function, first introduced by~\cite{hampel1974influence}, assesses the impact of individual data points on model performance and has become a powerful tool for training data valuation. Unlike alternative approaches such as LLM-based rating methods~\citep{liu2024a} or reward function methods~\citep{DBLP:journals/corr/abs-2410-06508}, the influence function offers distinct advantages by quantifying data utility through rigorous mathematical analysis of model training dynamics. 
Recent studies have extended its use to improve data quality in LLM pre-training through TraceIn~\citep{DBLP:conf/nips/PruthiLKS20} and MATES~\citep{yu2024mates}, for instruction tuning with Montessori-instruct~\citep{DBLP:journals/corr/abs-2410-14208} and LESS~\citep{DBLP:conf/icml/XiaMGA024}, and for reward modeling with OPORP~\citep{min2025understandingimpacthumanfeedback}. 
However, its potential for MAS data synthesis that maximizes system capability enhancement remains unexplored. The core challenge in applying influence functions lies in its high computational cost. Classical methods, such as gradient-based approaches~\citep{DBLP:conf/icml/KohL17, DBLP:conf/icml/ParkGILM23} and trajectory-influence based methods~\citep{DBLP:journals/corr/abs-2405-12186}, require the computation of billion-level gradients, which is extremely expensive. For efficient estimation, MATES~\citep{yu2024mates} probes the oracle data influence by evaluating the model's reference loss after training on individual data points. Our approach extends the reference loss to non-differentiable validation metrics, thereby enabling the enhancement of data quality through data synthesis.

%% file: table/efficiency_comparison.tex
\begin{table*}[t]
    \centering
    \begin{tabular}{lcc}
    \toprule
       \textbf{Methods} & \textbf{FLOPs} & \textbf{FLOPs ($10^{15}$)} \\
    \midrule
    DITS & $6N \cdot S + 2N \cdot V \cdot S$ & 0.7 \\
TRAK & $6N \cdot S \cdot V + 2N \cdot V \cdot d + V \cdot d^3$ & 4.6 \\
LESS & $6N \cdot S \cdot V \cdot R + 2N \cdot V \cdot d \cdot R$ & 18 \\
    \bottomrule
    \end{tabular}
    \caption{The computational costs comparison for estimating the influence score of one data point for different methods.}
    \label{tab:efficiency}
\end{table*}

%% file: table/selection_strategy.tex
\begin{table*}[t]
\centering
\begin{tabular}{lcc}
\toprule
Model & 2WMH QA (V=20) & 2WMH QA (V=5) \\
\midrule
Optima-DPO-Random-Select & 60.6 & 60.6 \\
DITS-DPO-High-Qvalue-High-Influence & 61.5 & 61.4 \\
DITS-DPO-Low-Qvalue-High-Influence & 61.7 & 60.8 \\
DITS-DPO-High-Qvalue-Low-Influence & 59.8 & 60.4 \\
\bottomrule
\end{tabular}
\caption{Performance comparison under different selection strategy.}
\label{tab:selection_strategy}
\end{table*}

%% file: table/robustness_table.tex
\begin{table*}[t]
\centering
\caption{Robustness of DITS-DPO under different validation sets on 2WMH QA.}
\label{tab:validation_robustness}
\begin{tabular}{lccccc}
\toprule
Model & Val Set 1 & Val Set 2 & Val Set 3 & Val Set 4 & Val Set 5 \\
\midrule
DITS-DPO & 62.2 & 61.8 & 62.0 & 62.4 & 61.9 \\
\bottomrule
\end{tabular}
\end{table*}

%% file: table/base_model.tex
\begin{table*}[t]
    \centering
    
    \resizebox{0.75\textwidth}{!}{%
    \begin{tabular}{lcccccc}
    \toprule
     \textbf{Method} & \multicolumn{1}{c}{\textbf{HotpotQA}} & \multicolumn{1}{c}{\textbf{2WMH QA}}  & \multicolumn{1}{c}{\textbf{CBT}}& \multicolumn{1}{c}{\textbf{ARC-C}}&\multicolumn{1}{c}{\textbf{MMLU}} \\
     \midrule
    Qwen2.5-7B-Instruct & & & & &  \\
    $\quad$- Base       & 39.02 & 37.92 & 25.77 & 4.93 & 11.40\\
    $\quad$- Optima-SFT & 49.08 & 58.18 & 52.63 & 73.98 & \textbf{57.70}\\
    $\quad$- Optima-DPO & \textbf{52.19} & \underline{60.86} & \underline{61.67} & \underline{74.05} & 55.70\\
    $\quad$- DITS-DPO   & \underline{51.68} & \textbf{61.81} & \textbf{62.23} & \textbf{74.23} & \underline{56.30} \\
    \midrule
    Llama-3.1-8B-Instruct & & &  & & \\
    $\quad$- Base       & 38.07 & 35.32 & 28.27 & 18.17 & 16.20 \\
    $\quad$- Optima-SFT & \underline{44.60} & 38.25 & 53.79 & 68.26 & 48.10 \\
    $\quad$- Optima-DPO & 44.47 & \underline{40.12} & \underline{54.29} & \underline{74.15} & \underline{56.20} \\
    $\quad$- DITS-DPO   & \textbf{45.23} & \textbf{41.22} & \textbf{61.44} & \textbf{76.37} & \textbf{58.00}\\
    \bottomrule
    \end{tabular}
    }
    \caption{\textbf{Results with different base models across Information Exchange and Debate tasks.} Best results are indicated in \textbf{bold}, and second-best results are \underline{underlined}.}
    \label{tab:base-model}
\end{table*}

%% file: table/hyper_parameter_main.tex
\begin{table*}[t]
    \centering
    \small
    \setlength{\tabcolsep}{3pt}
    \begin{tabular}{l*{7}{r}}
    \toprule
        & \textbf{HotpotQA} & \textbf{2WMH QA} & \textbf{TriviaQA} & \textbf{CBT} & \textbf{ARC-C} & \textbf{MMLU}\\
    \toprule
      \textit{\textbf{DITS-iSFT-DPO}} &  \\
      $\gamma$ & 1 & 1 & 1 & 1  & 1 & 1 \\
      $\alpha$ & 0.5 & 0.5 & 0.5 & 0.5 & 0.5 & 0.5 \\
      SFT LR & 2e-5 & 2e-5 & 2e-5 & 2e-5 & 1e-6 & 1e-6 \\
      SFT Epoch &  2  & 1 & 1 & 1 & 4 & 2\\
      SFT Batch Size& 32  & 32 & 32 & 32 & 16 & 16\\
      $\lambda_{token}$ & 0.6 & 0.6 & 0.6 & 0.6 & 0.5 & 0.6\\
      $\lambda_{loss}$& 1 & 1 & 1 & 1 & 0.6 & 0.7\\
      $\lambda_\text{dpo-filter}$ & 0.4 & 0.4 & 0.4 & 0.4 & 0.45 & 0.4\\
      $\lambda_\text{dpo-diff}$ & 0.2 & 0.2 & 0.2 & 0.2 & 0.2 & 0.2\\
      \textit{Iteration 0} & \\
      $\quad$ DPO LR & 5e-7 & 5e-7 & 5e-7 & 5e-7 & 5e-7 & 5e-7  \\
      $\quad$ DPO Epoch & 1  & 1 & 1 & 1 & 1 & 1\\
      $\quad$ DPO Batch Size& 64 & 64 & 64 & 64 & 64 & 64 \\
      $\quad$ $\beta$ & 0.5 & 0.5 & 0.7 & 0.7 & 0.1 & 0.1 \\
       \textit{Iteration 1} & \\
      $\quad$ DPO LR & 5e-7 & 5e-7 & 5e-7 & 5e-7 & 5e-7 & 5e-7  \\
      $\quad$ DPO Epoch & 1  & 1 & 1 & 1 & 1 & 1\\
      $\quad$ DPO Batch Size& 64 & 64 & 64 & 64 & 64 & 64 \\
      $\quad$ $\beta$ & 0.5 & 0.5 & 0.7 & 0.7 & 0.1 & 0.1 \\
       \textit{Iteration 2} & \\
      $\quad$ DPO LR & 5e-7 & 5e-7 & 5e-7 & 5e-7 & 5e-7 & 1e-6  \\
      $\quad$ DPO Epoch & 1  & 1 & 1 & 1 & 1 & 1\\
      $\quad$ DPO Batch Size& 64 & 64 & 64 & 64 & 64 & 64 \\
      $\quad$ $\beta$ & 0.5 & 0.5 & 0.7 & 0.5 & 0.2 & 0.1 \\
    \bottomrule
    \end{tabular}
    \caption{Hyper-parameters used in Table~\ref{tab:main-table}.}
    \label{tab:hyper_parameters_main}
\end{table*}

%% file: table/hyper_parameter_single.tex

\begin{table*}[t]
    \centering
    \small
    \setlength{\tabcolsep}{3pt}
    \begin{tabular}{l*{7}{r}}
    \toprule
        & \textbf{HotpotQA} & \textbf{2WMH QA} & \textbf{TriviaQA} & \textbf{CBT} & \textbf{ARC-C} & \textbf{MMLU}\\
    \toprule
      \textit{\textbf{DITS-DPO}} &  \\
      $\gamma$ & 1 & 1 & 1 & 1  & 1 & 1 \\
      $\alpha$ & 0.5 & 0.5 & 0.5 & 0.5 & 0.5 & 0.5 \\
      $\lambda_{token}$ & 0.6 & 0.6 & 0.6 & 0.6 & 0.5 & 0.6\\
      $\lambda_{loss}$& 1 & 1 & 1 & 1 & 0.6 & 0.7\\
      $\lambda_\text{dpo-filter}$ & 0.4 & 0.4 & 0.4 & 0.4 & 0.45 & 0.4\\
      $\lambda_\text{dpo-diff}$ & 0.2 & 0.2 & 0.2 & 0.2 & 0.2 & 0.2\\
      DPO LR & 5e-6 & 5e-7 & 5e-6 & 5e-7 & 5e-7 & 5e-7  \\
      DPO Epoch & 1  & 1 & 1 & 1 & 1 & 1\\
      DPO Batch Size& 64 & 64 & 64 & 64 & 64 & 64 \\
      $\beta$ & 0.5 & 0.5 & 0.7 & 0.5 & 0.4 & 0.1 \\
    \bottomrule
    \end{tabular}
    \caption{Hyper-parameters used in Table~\ref{tab:single-iteration}.}
    \label{tab:hyper_parameters_single}
\end{table*}

%% file: table/case.tex
\begin{table*}[t]

\centering
{\fontsize{8}{10}\selectfont
\begin{tabular}{|m{0.15\textwidth}|m{0.8\textwidth}|}
\toprule
    \textbf{Question} & Which film has the director who was born later, Eyes Of The Forest or Stardust On The Sage?\\
    \midrule
    \textbf{Q-value Select} & 
     \begin{Verbatim}[commandchars=\\\{\}]
Alice:
[
    [""Film"", ""Eyes of the Forest""],
    [""Director"", ""Lambert Hillyer""],
    [""Birth Date"", ""July 8, 1893""],
    [""Death Date"", ""July 5, 1969""],
    [""Director Born Later?"", ""0""]
]
Bob:
[
    [""Film"", ""Stardust On The Sage""],
    [""Director"", ""William Morgan""],
    [""Birth Date"", ""1895-12-07""],
    [""Director Born Later?"", ""1""]
]
\textcolor{blue}{Chosen}:
Alice:
[
    [""Director Born Later?"", ""1""],
    \textcolor{red}{[""Conclusion"", ""<A> Stardust On The Sage</A>""]}
]
\textcolor{blue}{Rejected}: 
Alice:
[
    [""Director"", ""Lambert Hillyer""],
    [""Birth Date"", ""July 8, 1893""],
    \textcolor{red}{[""Director Born Later?"", ""0""]}
]
     \end{Verbatim}
\\
\midrule
\textbf{Influence Score Select} &

\begin{Verbatim}[commandchars=\\\{\}]
Alice:
[
    [""Information"", ""Lambert Harwood Hillyer was born 1893""],
    [""Query"", ""Is Hillyer the director of Eyes Of The Forest?""],
    [""Information"", ""Yes""]
]
Bob:
[
    [""Query"", ""Is Hillyer the director of Stardust On The Sage?""],
    [""Information"", ""No""]
]
\textcolor{blue}{Chosen}:
Alice:
[
    \textcolor{red}{[""Query"", ""Is Hillyer the director of Stardust On The Sage?""],}
    \textcolor{red}{[""Information"", ""No""]}
]
\textcolor{blue}{Rejected}:
Alice:
[
    [""Query"", ""Who is the director of Stardust On The Sage?""],
    [""Information"", ""Rumbi Katedza""],
    \textcolor{red}{[""Query"", ""Is Katedza the director of Stardust On The Sage?""],}
    \textcolor{red}{[""Information"", ""Yes""]}
]
\end{Verbatim}

    \\\bottomrule
\end{tabular}
}
\caption{
Case study to demonstrate the data selected by Q-value and influence score on 2WMH QA dataset.
}
\label{tab:cases}
\end{table*}

%% file: section/Impact.tex

%% file: main.bbl
\begin{thebibliography}{69}
\providecommand{\natexlab}[1]{#1}

\bibitem[{Bae et~al.(2024)Bae, Lin, Lorraine, and Grosse}]{DBLP:journals/corr/abs-2405-12186}
Juhan Bae, Wu~Lin, Jonathan Lorraine, and Roger Grosse. 2024.
\newblock Training data attribution via approximate unrolled differentiation.
\newblock \emph{CoRR}, abs/2405.12186.

\bibitem[{Bhakthavatsalam et~al.(2021)Bhakthavatsalam, Khashabi, Khot, Mishra, Richardson, Sabharwal, Schoenick, Tafjord, and Clark}]{DBLP:journals/corr/abs-2102-03315}
Sumithra Bhakthavatsalam, Daniel Khashabi, Tushar Khot, Bhavana~Dalvi Mishra, Kyle Richardson, Ashish Sabharwal, Carissa Schoenick, Oyvind Tafjord, and Peter Clark. 2021.
\newblock Think you have solved direct-answer question answering? try arc-da, the direct-answer {AI2} reasoning challenge.
\newblock \emph{CoRR}, abs/2102.03315.

\bibitem[{Bondy and Murty(1976)}]{Bondy1976}
J.~A. Bondy and U.~S.~R. Murty. 1976.
\newblock \emph{Graph Theory with Applications}.
\newblock Elsevier, New York.

\bibitem[{Chen et~al.(2023)Chen, Shu, Shareghi, Collier, Narasimhan, and Yao}]{DBLP:journals/corr/abs-2310-05915}
Baian Chen, Chang Shu, Ehsan Shareghi, Nigel Collier, Karthik Narasimhan, and Shunyu Yao. 2023.
\newblock Fireact: Toward language agent fine-tuning.
\newblock \emph{CoRR}, abs/2310.05915.

\bibitem[{Chen et~al.(2024{\natexlab{a}})Chen, Yuan, Yuan, Su, Qian, Yang, Xie, Liu, and Sun}]{DBLP:conf/emnlp/ChenYYSQYXL024}
Weize Chen, Chenfei Yuan, Jiarui Yuan, Yusheng Su, Chen Qian, Cheng Yang, Ruobing Xie, Zhiyuan Liu, and Maosong Sun. 2024{\natexlab{a}}.
\newblock Beyond natural language: Llms leveraging alternative formats for enhanced reasoning and communication.
\newblock In \emph{{EMNLP} (Findings)}, pages 10626--10641. Association for Computational Linguistics.

\bibitem[{Chen et~al.(2024{\natexlab{b}})Chen, Yuan, Qian, Yang, Liu, and Sun}]{DBLP:journals/corr/abs-2410-08115}
Weize Chen, Jiarui Yuan, Chen Qian, Cheng Yang, Zhiyuan Liu, and Maosong Sun. 2024{\natexlab{b}}.
\newblock Optima: Optimizing effectiveness and efficiency for llm-based multi-agent system.
\newblock \emph{CoRR}, abs/2410.08115.

\bibitem[{Du et~al.(2024)Du, Li, Torralba, Tenenbaum, and Mordatch}]{DBLP:conf/icml/Du00TM24}
Yilun Du, Shuang Li, Antonio Torralba, Joshua~B. Tenenbaum, and Igor Mordatch. 2024.
\newblock Improving factuality and reasoning in language models through multiagent debate.
\newblock In \emph{{ICML}}. OpenReview.net.

\bibitem[{Dubey et~al.(2024)Dubey, Jauhri, and et~al}]{dubey2024llama}
Abhimanyu Dubey, Abhinav Jauhri, and et~al. 2024.
\newblock \href {https://arxiv.org/abs/2407.21783} {The llama 3 herd of models}.
\newblock \emph{Preprint}, arXiv:2407.21783.

\bibitem[{Gross and Yellen(2005)}]{book:gross:2005}
Jonathan~L. Gross and Jay Yellen. 2005.
\newblock \href {http://books.google.com/books?vid=ISBN158488505X} {\emph{Graph Theory and Its Applications}}.

\bibitem[{Guan et~al.(2025)Guan, Zhang, Liu, Shang, Sun, Zhu, Yang, and Yang}]{guan2025rstarmathsmallllmsmaster}
Xinyu Guan, Li~Lyna Zhang, Yifei Liu, Ning Shang, Youran Sun, Yi~Zhu, Fan Yang, and Mao Yang. 2025.
\newblock \href {https://arxiv.org/abs/2501.04519} {rstar-math: Small llms can master math reasoning with self-evolved deep thinking}.
\newblock \emph{Preprint}, arXiv:2501.04519.

\bibitem[{Guo et~al.(2024)Guo, Chen, Wang, Chang, Pei, Chawla, Wiest, and Zhang}]{DBLP:conf/ijcai/GuoCWCPCW024}
Taicheng Guo, Xiuying Chen, Yaqi Wang, Ruidi Chang, Shichao Pei, Nitesh~V. Chawla, Olaf Wiest, and Xiangliang Zhang. 2024.
\newblock Large language model based multi-agents: {A} survey of progress and challenges.
\newblock In \emph{{IJCAI}}, pages 8048--8057. ijcai.org.

\bibitem[{Hampel(1974)}]{hampel1974influence}
Frank~R Hampel. 1974.
\newblock The influence curve and its role in robust estimation.
\newblock \emph{Journal of the american statistical association}, 69(346):383--393.

\bibitem[{Hao et~al.(2023)Hao, Gu, Ma, Hong, Wang, Wang, and Hu}]{DBLP:conf/emnlp/HaoGMHWWH23}
Shibo Hao, Yi~Gu, Haodi Ma, Joshua~Jiahua Hong, Zhen Wang, Daisy~Zhe Wang, and Zhiting Hu. 2023.
\newblock Reasoning with language model is planning with world model.
\newblock In \emph{{EMNLP}}, pages 8154--8173. Association for Computational Linguistics.

\bibitem[{Hendrycks et~al.(2021)Hendrycks, Burns, Basart, Zou, Mazeika, Song, and Steinhardt}]{DBLP:conf/iclr/HendrycksBBZMSS21}
Dan Hendrycks, Collin Burns, Steven Basart, Andy Zou, Mantas Mazeika, Dawn Song, and Jacob Steinhardt. 2021.
\newblock Measuring massive multitask language understanding.
\newblock In \emph{{ICLR}}. OpenReview.net.

\bibitem[{Hill et~al.(2016)Hill, Bordes, Chopra, and Weston}]{DBLP:journals/corr/HillBCW15}
Felix Hill, Antoine Bordes, Sumit Chopra, and Jason Weston. 2016.
\newblock The goldilocks principle: Reading children's books with explicit memory representations.
\newblock In \emph{{ICLR}}.

\bibitem[{Ho et~al.(2020)Ho, Nguyen, Sugawara, and Aizawa}]{DBLP:conf/coling/HoNSA20}
Xanh Ho, Anh{-}Khoa~Duong Nguyen, Saku Sugawara, and Akiko Aizawa. 2020.
\newblock Constructing {A} multi-hop {QA} dataset for comprehensive evaluation of reasoning steps.
\newblock In \emph{{COLING}}, pages 6609--6625. International Committee on Computational Linguistics.

\bibitem[{Hong et~al.(2024)Hong, Zhuge, Chen, Zheng, Cheng, Wang, Zhang, Wang, Yau, Lin, Zhou, Ran, Xiao, Wu, and Schmidhuber}]{DBLP:conf/iclr/HongZCZCWZWYLZR24}
Sirui Hong, Mingchen Zhuge, Jonathan Chen, Xiawu Zheng, Yuheng Cheng, Jinlin Wang, Ceyao Zhang, Zili Wang, Steven Ka~Shing Yau, Zijuan Lin, Liyang Zhou, Chenyu Ran, Lingfeng Xiao, Chenglin Wu, and J{\"{u}}rgen Schmidhuber. 2024.
\newblock Metagpt: Meta programming for {A} multi-agent collaborative framework.
\newblock In \emph{{ICLR}}. OpenReview.net.

\bibitem[{Hu et~al.(2022)Hu, Shen, Wallis, Allen{-}Zhu, Li, Wang, Wang, and Chen}]{DBLP:conf/iclr/HuSWALWWC22}
Edward~J. Hu, Yelong Shen, Phillip Wallis, Zeyuan Allen{-}Zhu, Yuanzhi Li, Shean Wang, Lu~Wang, and Weizhu Chen. 2022.
\newblock Lora: Low-rank adaptation of large language models.
\newblock In \emph{{ICLR}}. OpenReview.net.

\bibitem[{Hu et~al.(2024)Hu, Xiong, Yi, Wei, Xiao, Chen, Ye, Tao, Zhou, Zhao, Li, Xu, Wang, Xu, Qiao, Kuang, Zeng, Wang, Li, Jiang, Zhou, Wang, Yin, Zhao, Yang, Wu, Zhang, and Wu}]{202412.2294}
Xueyu Hu, Tao Xiong, Biao Yi, Zishu Wei, Ruixuan Xiao, Yurun Chen, Jiasheng Ye, Meiling Tao, Xiangxin Zhou, Ziyu Zhao, Yuhuai Li, Shengze Xu, Shawn Wang, Xinchen Xu, Shuofei Qiao, Kun Kuang, Tieyong Zeng, Liang Wang, Jiwei Li, and 9 others. 2024.
\newblock Os agents: A survey on mllm-based agents for general computing devices use.
\newblock \emph{Preprints}.

\bibitem[{Islam et~al.(2024)Islam, Ali, and Parvez}]{DBLP:conf/acl/IslamAP24}
Md.~Ashraful Islam, Mohammed~Eunus Ali, and Md.~Rizwan Parvez. 2024.
\newblock Mapcoder: Multi-agent code generation for competitive problem solving.
\newblock In \emph{{ACL} {(1)}}, pages 4912--4944. Association for Computational Linguistics.

\bibitem[{Joshi et~al.(2017)Joshi, Choi, Weld, and Zettlemoyer}]{DBLP:conf/acl/JoshiCWZ17}
Mandar Joshi, Eunsol Choi, Daniel~S. Weld, and Luke Zettlemoyer. 2017.
\newblock Triviaqa: {A} large scale distantly supervised challenge dataset for reading comprehension.
\newblock In \emph{{ACL} {(1)}}, pages 1601--1611. Association for Computational Linguistics.

\bibitem[{Khattab et~al.(2023)Khattab, Singhvi, Maheshwari, Zhang, Santhanam, Vardhamanan, Haq, Sharma, Joshi, Moazam, Miller, Zaharia, and Potts}]{DBLP:journals/corr/abs-2310-03714}
Omar Khattab, Arnav Singhvi, Paridhi Maheshwari, Zhiyuan Zhang, Keshav Santhanam, Sri Vardhamanan, Saiful Haq, Ashutosh Sharma, Thomas~T. Joshi, Hanna Moazam, Heather Miller, Matei Zaharia, and Christopher Potts. 2023.
\newblock Dspy: Compiling declarative language model calls into self-improving pipelines.
\newblock \emph{CoRR}, abs/2310.03714.

\bibitem[{Koh and Liang(2017)}]{DBLP:conf/icml/KohL17}
Pang~Wei Koh and Percy Liang. 2017.
\newblock Understanding black-box predictions via influence functions.
\newblock In \emph{{ICML}}, volume~70 of \emph{Proceedings of Machine Learning Research}, pages 1885--1894. {PMLR}.

\bibitem[{Li et~al.(2023)Li, Hammoud, Itani, Khizbullin, and Ghanem}]{DBLP:conf/nips/LiHIKG23}
Guohao Li, Hasan Hammoud, Hani Itani, Dmitrii Khizbullin, and Bernard Ghanem. 2023.
\newblock {CAMEL:} communicative agents for "mind" exploration of large language model society.
\newblock In \emph{NeurIPS}.

\bibitem[{Li et~al.(2024{\natexlab{a}})Li, Zhang, Yu, Fu, and Ye}]{DBLP:journals/corr/abs-2402-05120}
Junyou Li, Qin Zhang, Yangbin Yu, Qiang Fu, and Deheng Ye. 2024{\natexlab{a}}.
\newblock More agents is all you need.
\newblock \emph{CoRR}, abs/2402.05120.

\bibitem[{Li et~al.(2025{\natexlab{a}})Li, Dong, Luan, Di, and Ding}]{li2025enhancingreasoningprocesssupervision}
Shuangtao Li, Shuaihao Dong, Kexin Luan, Xinhan Di, and Chaofan Ding. 2025{\natexlab{a}}.
\newblock \href {https://arxiv.org/abs/2501.01478} {Enhancing reasoning through process supervision with monte carlo tree search}.
\newblock \emph{Preprint}, arXiv:2501.01478.

\bibitem[{Li et~al.(2024{\natexlab{b}})Li, Yu, and Xiong}]{DBLP:journals/corr/abs-2410-14208}
Xiaochuan Li, Zichun Yu, and Chenyan Xiong. 2024{\natexlab{b}}.
\newblock Montessori-instruct: Generate influential training data tailored for student learning.
\newblock \emph{CoRR}, abs/2410.14208.

\bibitem[{Li et~al.(2025{\natexlab{b}})Li, Dong, Jin, Zhang, Zhou, Zhu, Zhang, and Dou}]{DBLP:journals/corr/abs-2501-05366}
Xiaoxi Li, Guanting Dong, Jiajie Jin, Yuyao Zhang, Yujia Zhou, Yutao Zhu, Peitian Zhang, and Zhicheng Dou. 2025{\natexlab{b}}.
\newblock \href {https://doi.org/10.48550/ARXIV.2501.05366} {Search-o1: Agentic search-enhanced large reasoning models}.
\newblock \emph{CoRR}, abs/2501.05366.

\bibitem[{Li et~al.(2025{\natexlab{c}})Li, Jin, Dong, Qian, Zhu, Wu, Wen, and Dou}]{Li2025WebThinker}
Xiaoxi Li, Jiajie Jin, Guanting Dong, Hongjin Qian, Yutao Zhu, Yongkang Wu, Ji{-}Rong Wen, and Zhicheng Dou. 2025{\natexlab{c}}.
\newblock \href {https://doi.org/10.48550/ARXIV.2504.21776} {Webthinker: Empowering large reasoning models with deep research capability}.
\newblock \emph{CoRR}, abs/2504.21776.

\bibitem[{Liang et~al.(2024)Liang, He, Jiao, Wang, Wang, Wang, Yang, Shi, and Tu}]{DBLP:conf/emnlp/Liang0JW00Y0T24}
Tian Liang, Zhiwei He, Wenxiang Jiao, Xing Wang, Yan Wang, Rui Wang, Yujiu Yang, Shuming Shi, and Zhaopeng Tu. 2024.
\newblock Encouraging divergent thinking in large language models through multi-agent debate.
\newblock In \emph{{EMNLP}}, pages 17889--17904. Association for Computational Linguistics.

\bibitem[{Liu et~al.(2024)Liu, Zhang, Li, Liu, and Yang}]{liu2024a}
Zijun Liu, Yanzhe Zhang, Peng Li, Yang Liu, and Diyi Yang. 2024.
\newblock A dynamic {LLM}-powered agent network for task-oriented agent collaboration.
\newblock In \emph{COLM}.

\bibitem[{Min et~al.(2025)Min, Lee, Ryu, Kwon, and Lee}]{min2025understandingimpacthumanfeedback}
Taywon Min, Haeone Lee, Hanho Ryu, Yongchan Kwon, and Kimin Lee. 2025.
\newblock \href {https://arxiv.org/abs/2501.05790} {Understanding impact of human feedback via influence functions}.
\newblock \emph{Preprint}, arXiv:2501.05790.

\bibitem[{Motwani et~al.(2024)Motwani, Smith, Das, Rybchuk, Torr, Laptev, Pizzati, Clark, and de~Witt}]{DBLP:journals/corr/abs-2412-01928}
Sumeet~Ramesh Motwani, Chandler Smith, Rocktim~Jyoti Das, Markian Rybchuk, Philip H.~S. Torr, Ivan Laptev, Fabio Pizzati, Ronald Clark, and Christian~Schr{\"{o}}der de~Witt. 2024.
\newblock {MALT:} improving reasoning with multi-agent {LLM} training.
\newblock \emph{CoRR}, abs/2412.01928.

\bibitem[{OpenAI(2023)}]{DBLP:journals/corr/abs-2303-08774}
OpenAI. 2023.
\newblock {GPT-4} technical report.
\newblock \emph{CoRR}, abs/2303.08774.

\bibitem[{Opsahl{-}Ong et~al.(2024)Opsahl{-}Ong, Ryan, Purtell, Broman, Potts, Zaharia, and Khattab}]{DBLP:conf/emnlp/Opsahl-OngRPBPZ24}
Krista Opsahl{-}Ong, Michael~J. Ryan, Josh Purtell, David Broman, Christopher Potts, Matei Zaharia, and Omar Khattab. 2024.
\newblock Optimizing instructions and demonstrations for multi-stage language model programs.
\newblock In \emph{{EMNLP}}, pages 9340--9366. Association for Computational Linguistics.

\bibitem[{Pang et~al.(2024)Pang, Yuan, Cho, He, Sukhbaatar, and Weston}]{DBLP:journals/corr/abs-2404-19733}
Richard~Yuanzhe Pang, Weizhe Yuan, Kyunghyun Cho, He~He, Sainbayar Sukhbaatar, and Jason Weston. 2024.
\newblock Iterative reasoning preference optimization.
\newblock \emph{CoRR}, abs/2404.19733.

\bibitem[{Park et~al.(2023)Park, Georgiev, Ilyas, Leclerc, and Madry}]{DBLP:conf/icml/ParkGILM23}
Sung~Min Park, Kristian Georgiev, Andrew Ilyas, Guillaume Leclerc, and Aleksander Madry. 2023.
\newblock {TRAK:} attributing model behavior at scale.
\newblock In \emph{{ICML}}, volume 202 of \emph{Proceedings of Machine Learning Research}, pages 27074--27113. {PMLR}.

\bibitem[{Pruthi et~al.(2020)Pruthi, Liu, Kale, and Sundararajan}]{DBLP:conf/nips/PruthiLKS20}
Garima Pruthi, Frederick Liu, Satyen Kale, and Mukund Sundararajan. 2020.
\newblock Estimating training data influence by tracing gradient descent.
\newblock In \emph{NeurIPS}.

\bibitem[{Qi et~al.(2024)Qi, Ma, Xu, Zhang, Yang, and Yang}]{DBLP:journals/corr/abs-2408-06195}
Zhenting Qi, Mingyuan Ma, Jiahang Xu, Li~Lyna Zhang, Fan Yang, and Mao Yang. 2024.
\newblock Mutual reasoning makes smaller llms stronger problem-solvers.
\newblock \emph{CoRR}, abs/2408.06195.

\bibitem[{Qian et~al.(2024{\natexlab{a}})Qian, Li, Dang, Liu, Wang, Xie, Chen, Yang, Zhang, Liu, and Sun}]{DBLP:journals/corr/abs-2405-04219}
Chen Qian, Jiahao Li, Yufan Dang, Wei Liu, Yifei Wang, Zihao Xie, Weize Chen, Cheng Yang, Yingli Zhang, Zhiyuan Liu, and Maosong Sun. 2024{\natexlab{a}}.
\newblock Iterative experience refinement of software-developing agents.
\newblock \emph{CoRR}, abs/2405.04219.

\bibitem[{Qian et~al.(2024{\natexlab{b}})Qian, Liu, Liu, Chen, Dang, Li, Yang, Chen, Su, Cong, Xu, Li, Liu, and Sun}]{DBLP:conf/acl/QianLLCDL0CSCXL24}
Chen Qian, Wei Liu, Hongzhang Liu, Nuo Chen, Yufan Dang, Jiahao Li, Cheng Yang, Weize Chen, Yusheng Su, Xin Cong, Juyuan Xu, Dahai Li, Zhiyuan Liu, and Maosong Sun. 2024{\natexlab{b}}.
\newblock Chatdev: Communicative agents for software development.
\newblock In \emph{{ACL} {(1)}}, pages 15174--15186. Association for Computational Linguistics.

\bibitem[{Qian et~al.(2024{\natexlab{c}})Qian, Xie, Wang, Liu, Dang, Du, Chen, Yang, Liu, and Sun}]{qian2024}
Chen Qian, Zihao Xie, Yifei Wang, Wei Liu, Yufan Dang, Zhuoyun Du, Weize Chen, Cheng Yang, Zhiyuan Liu, and Maosong Sun. 2024{\natexlab{c}}.
\newblock \href {https://arxiv.org/abs/2406.07155} {Scaling large-language-model-based multi-agent collaboration}.
\newblock \emph{Preprint}, arXiv:2406.07155.

\bibitem[{Rafailov et~al.(2024)Rafailov, Chittepu, Park, Sikchi, Hejna, Knox, Finn, and Niekum}]{DBLP:journals/corr/abs-2406-02900}
Rafael Rafailov, Yaswanth Chittepu, Ryan Park, Harshit Sikchi, Joey Hejna, W.~Bradley Knox, Chelsea Finn, and Scott Niekum. 2024.
\newblock Scaling laws for reward model overoptimization in direct alignment algorithms.
\newblock \emph{CoRR}, abs/2406.02900.

\bibitem[{Rafailov et~al.(2023)Rafailov, Sharma, Mitchell, Manning, Ermon, and Finn}]{DBLP:conf/nips/RafailovSMMEF23}
Rafael Rafailov, Archit Sharma, Eric Mitchell, Christopher~D. Manning, Stefano Ermon, and Chelsea Finn. 2023.
\newblock Direct preference optimization: Your language model is secretly a reward model.
\newblock In \emph{NeurIPS}.

\bibitem[{Shi et~al.(2024{\natexlab{a}})Shi, He, Zhang, Gao, Li, Zhang, Wang, and Feng}]{DBLP:conf/sigir/Shi0ZGLZWF24}
Wentao Shi, Xiangnan He, Yang Zhang, Chongming Gao, Xinyue Li, Jizhi Zhang, Qifan Wang, and Fuli Feng. 2024{\natexlab{a}}.
\newblock Large language models are learnable planners for long-term recommendation.
\newblock In \emph{{SIGIR}}, pages 1893--1903. {ACM}.

\bibitem[{Shi et~al.(2024{\natexlab{b}})Shi, Yuan, Wu, Wang, and Feng}]{DBLP:conf/emnlp/ShiYWWF24}
Wentao Shi, Mengqi Yuan, Junkang Wu, Qifan Wang, and Fuli Feng. 2024{\natexlab{b}}.
\newblock Direct multi-turn preference optimization for language agents.
\newblock In \emph{{EMNLP}}, pages 2312--2324. Association for Computational Linguistics.

\bibitem[{Shi et~al.(2024{\natexlab{c}})Shi, Land, Locatelli, Geist, and Bartolo}]{DBLP:journals/corr/abs-2410-11677}
Zhengyan Shi, Sander Land, Acyr Locatelli, Matthieu Geist, and Max Bartolo. 2024{\natexlab{c}}.
\newblock Understanding likelihood over-optimisation in direct alignment algorithms.
\newblock \emph{CoRR}, abs/2410.11677.

\bibitem[{Silver et~al.(2016)Silver, Huang, Maddison, Guez, Sifre, van~den Driessche, Schrittwieser, Antonoglou, Panneershelvam, Lanctot, Dieleman, Grewe, Nham, Kalchbrenner, Sutskever, Lillicrap, Leach, Kavukcuoglu, Graepel, and Hassabis}]{DBLP:journals/nature/SilverHMGSDSAPL16}
David Silver, Aja Huang, Chris~J. Maddison, Arthur Guez, Laurent Sifre, George van~den Driessche, Julian Schrittwieser, Ioannis Antonoglou, Vedavyas Panneershelvam, Marc Lanctot, Sander Dieleman, Dominik Grewe, John Nham, Nal Kalchbrenner, Ilya Sutskever, Timothy~P. Lillicrap, Madeleine Leach, Koray Kavukcuoglu, Thore Graepel, and Demis Hassabis. 2016.
\newblock Mastering the game of go with deep neural networks and tree search.
\newblock \emph{Nat.}, 529(7587):484--489.

\bibitem[{Snell et~al.(2024)Snell, Lee, Xu, and Kumar}]{DBLP:journals/corr/abs-2408-03314}
Charlie Snell, Jaehoon Lee, Kelvin Xu, and Aviral Kumar. 2024.
\newblock Scaling {LLM} test-time compute optimally can be more effective than scaling model parameters.
\newblock \emph{CoRR}, abs/2408.03314.

\bibitem[{Team(2024)}]{QwenTeam2024Qwq}
Qwen Team. 2024.
\newblock Qwq: Reflect deeply on the boundaries of the unknown.
\newblock \url{https://huggingface.co}.
\newblock [50].

\bibitem[{Team(2025)}]{qwen2025qwen25technicalreport}
Qwen Team. 2025.
\newblock \href {https://arxiv.org/abs/2412.15115} {Qwen2.5 technical report}.
\newblock \emph{Preprint}, arXiv:2412.15115.

\bibitem[{Tran et~al.(2025)Tran, Dao, Nguyen, Pham, O'Sullivan, and Nguyen}]{tran2025multiagentcollaborationmechanismssurvey}
Khanh-Tung Tran, Dung Dao, Minh-Duong Nguyen, Quoc-Viet Pham, Barry O'Sullivan, and Hoang~D. Nguyen. 2025.
\newblock \href {https://arxiv.org/abs/2501.06322} {Multi-agent collaboration mechanisms: A survey of llms}.
\newblock \emph{Preprint}, arXiv:2501.06322.

\bibitem[{Wang et~al.(2024{\natexlab{a}})Wang, Wang, Athiwaratkun, Zhang, and Zou}]{DBLP:journals/corr/abs-2406-04692}
Junlin Wang, Jue Wang, Ben Athiwaratkun, Ce~Zhang, and James Zou. 2024{\natexlab{a}}.
\newblock Mixture-of-agents enhances large language model capabilities.
\newblock \emph{CoRR}, abs/2406.04692.

\bibitem[{Wang et~al.(2024{\natexlab{b}})Wang, Ma, Feng, Zhang, Yang, Zhang, Chen, Tang, Chen, Lin, Zhao, Wei, and Wen}]{Wang_2024}
Lei Wang, Chen Ma, Xueyang Feng, Zeyu Zhang, Hao Yang, Jingsen Zhang, Zhiyuan Chen, Jiakai Tang, Xu~Chen, Yankai Lin, Wayne~Xin Zhao, Zhewei Wei, and Jirong Wen. 2024{\natexlab{b}}.
\newblock A survey on large language model based autonomous agents.
\newblock \emph{Frontiers of Computer Science}, 18(6).

\bibitem[{Wang et~al.(2024{\natexlab{c}})Wang, Song, Tian, Yu, Peng, Mi, Huang, and Yu}]{DBLP:journals/corr/abs-2410-06508}
Xiyao Wang, Linfeng Song, Ye~Tian, Dian Yu, Baolin Peng, Haitao Mi, Furong Huang, and Dong Yu. 2024{\natexlab{c}}.
\newblock Towards self-improvement of llms via {MCTS:} leveraging stepwise knowledge with curriculum preference learning.
\newblock \emph{CoRR}, abs/2410.06508.

\bibitem[{Wang et~al.(2024{\natexlab{d}})Wang, Mao, Wu, Ge, Wei, and Ji}]{DBLP:conf/naacl/WangMW0WJ24}
Zhenhailong Wang, Shaoguang Mao, Wenshan Wu, Tao Ge, Furu Wei, and Heng Ji. 2024{\natexlab{d}}.
\newblock Unleashing the emergent cognitive synergy in large language models: {A} task-solving agent through multi-persona self-collaboration.
\newblock In \emph{{NAACL-HLT}}, pages 257--279. Association for Computational Linguistics.

\bibitem[{Wei et~al.(2022)Wei, Wang, Schuurmans, Bosma, Ichter, Xia, Chi, Le, and Zhou}]{DBLP:conf/nips/Wei0SBIXCLZ22}
Jason Wei, Xuezhi Wang, Dale Schuurmans, Maarten Bosma, Brian Ichter, Fei Xia, Ed~H. Chi, Quoc~V. Le, and Denny Zhou. 2022.
\newblock Chain-of-thought prompting elicits reasoning in large language models.
\newblock In \emph{NeurIPS}.

\bibitem[{Wu et~al.(2025)Wu, Yin, Jiang, Wang, Xi, Fang, Zhou, Xie, and Huang}]{wu2025webwalker}
Jialong Wu, Wenbiao Yin, Yong Jiang, Zhenglin Wang, Zekun Xi, Runnan Fang, Deyu Zhou, Pengjun Xie, and Fei Huang. 2025.
\newblock \href {https://arxiv.org/abs/2501.07572} {Webwalker: Benchmarking llms in web traversal}.
\newblock \emph{Preprint}, arXiv:2501.07572.

\bibitem[{Wu et~al.(2023)Wu, Bansal, Zhang, Wu, Zhang, Zhu, Li, Jiang, Zhang, and Wang}]{DBLP:journals/corr/abs-2308-08155}
Qingyun Wu, Gagan Bansal, Jieyu Zhang, Yiran Wu, Shaokun Zhang, Erkang Zhu, Beibin Li, Li~Jiang, Xiaoyun Zhang, and Chi Wang. 2023.
\newblock Autogen: Enabling next-gen {LLM} applications via multi-agent conversation framework.
\newblock \emph{CoRR}, abs/2308.08155.

\bibitem[{Wu et~al.(2024)Wu, Sun, Li, Welleck, and Yang}]{wu2024inferencescalinglawsempirical}
Yangzhen Wu, Zhiqing Sun, Shanda Li, Sean Welleck, and Yiming Yang. 2024.
\newblock \href {https://arxiv.org/abs/2408.00724} {Inference scaling laws: An empirical analysis of compute-optimal inference for problem-solving with language models}.
\newblock \emph{Preprint}, arXiv:2408.00724.

\bibitem[{Xi et~al.(2023)Xi, Chen, Guo, He, Ding, Hong, Zhang, Wang, Jin, Zhou, Zheng, Fan, Wang, Xiong, Zhou, Wang, Jiang, Zou, Liu, Yin, Dou, Weng, Cheng, Zhang, Qin, Zheng, Qiu, Huang, and Gui}]{xi2023risepotentiallargelanguage}
Zhiheng Xi, Wenxiang Chen, Xin Guo, Wei He, Yiwen Ding, Boyang Hong, Ming Zhang, Junzhe Wang, Senjie Jin, Enyu Zhou, Rui Zheng, Xiaoran Fan, Xiao Wang, Limao Xiong, Yuhao Zhou, Weiran Wang, Changhao Jiang, Yicheng Zou, Xiangyang Liu, and 10 others. 2023.
\newblock The rise and potential of large language model based agents: A survey.

\bibitem[{Xia et~al.(2024)Xia, Malladi, Gururangan, Arora, and Chen}]{DBLP:conf/icml/XiaMGA024}
Mengzhou Xia, Sadhika Malladi, Suchin Gururangan, Sanjeev Arora, and Danqi Chen. 2024.
\newblock {LESS:} selecting influential data for targeted instruction tuning.
\newblock In \emph{{ICML}}. OpenReview.net.

\bibitem[{Xie et~al.(2024)Xie, Goyal, Zheng, Kan, Lillicrap, Kawaguchi, and Shieh}]{DBLP:journals/corr/abs-2405-00451}
Yuxi Xie, Anirudh Goyal, Wenyue Zheng, Min{-}Yen Kan, Timothy~P. Lillicrap, Kenji Kawaguchi, and Michael Shieh. 2024.
\newblock Monte carlo tree search boosts reasoning via iterative preference learning.
\newblock \emph{CoRR}, abs/2405.00451.

\bibitem[{Yang et~al.(2018)Yang, Qi, Zhang, Bengio, Cohen, Salakhutdinov, and Manning}]{DBLP:conf/emnlp/Yang0ZBCSM18}
Zhilin Yang, Peng Qi, Saizheng Zhang, Yoshua Bengio, William~W. Cohen, Ruslan Salakhutdinov, and Christopher~D. Manning. 2018.
\newblock Hotpotqa: {A} dataset for diverse, explainable multi-hop question answering.
\newblock In \emph{{EMNLP}}, pages 2369--2380. Association for Computational Linguistics.

\bibitem[{Yao et~al.(2023)Yao, Zhao, Yu, Du, Shafran, Narasimhan, and Cao}]{DBLP:conf/iclr/YaoZYDSN023}
Shunyu Yao, Jeffrey Zhao, Dian Yu, Nan Du, Izhak Shafran, Karthik~R. Narasimhan, and Yuan Cao. 2023.
\newblock React: Synergizing reasoning and acting in language models.
\newblock In \emph{{ICLR}}. OpenReview.net.

\bibitem[{Yu et~al.(2024)Yu, Das, and Xiong}]{yu2024mates}
Zichun Yu, Spandan Das, and Chenyan Xiong. 2024.
\newblock Mates: Model-aware data selection for efficient pretraining with data influence models.
\newblock In \emph{NeurIPS}.

\bibitem[{Zhang et~al.(2024{\natexlab{a}})Zhang, He, Qian, Li, Li, Qin, Kang, Ma, Liu, Lin, Rajmohan, Zhang, and Zhang}]{zhang2024largelanguagemodelbrainedgui}
Chaoyun Zhang, Shilin He, Jiaxu Qian, Bowen Li, Liqun Li, Si~Qin, Yu~Kang, Minghua Ma, Guyue Liu, Qingwei Lin, Saravan Rajmohan, Dongmei Zhang, and Qi~Zhang. 2024{\natexlab{a}}.
\newblock \href {https://arxiv.org/abs/2411.18279} {Large language model-brained gui agents: A survey}.
\newblock \emph{Preprint}, arXiv:2411.18279.

\bibitem[{Zhang et~al.(2024{\natexlab{b}})Zhang, Zhoubian, Yue, Dong, and Tang}]{DBLP:journals/corr/abs-2406-03816}
Dan Zhang, Sining Zhoubian, Yisong Yue, Yuxiao Dong, and Jie Tang. 2024{\natexlab{b}}.
\newblock Rest-mcts*: {LLM} self-training via process reward guided tree search.
\newblock \emph{CoRR}, abs/2406.03816.

\bibitem[{Zhang et~al.(2024{\natexlab{c}})Zhang, Huang, Zhou, Li, and Ouyang}]{DBLP:journals/corr/abs-2406-07394}
Di~Zhang, Xiaoshui Huang, Dongzhan Zhou, Yuqiang Li, and Wanli Ouyang. 2024{\natexlab{c}}.
\newblock Accessing {GPT-4} level mathematical olympiad solutions via monte carlo tree self-refine with llama-3 8b.
\newblock \emph{CoRR}, abs/2406.07394.

\end{thebibliography}
